\documentclass[10pt,twocolumn,letterpaper]{article}

\usepackage{iccv}
\usepackage{times}
\usepackage{epsfig}
\usepackage{graphicx}
\usepackage{amsmath}
\usepackage{amssymb}
\usepackage{multirow}
\usepackage{makecell}
\usepackage{bbm}
\usepackage{commath}
\usepackage{lipsum}
\usepackage{stfloats}
\usepackage{comment}
\usepackage{caption}
\usepackage[dvipsnames]{xcolor}
\usepackage[pagebackref=true,breaklinks=true,letterpaper=true,colorlinks,bookmarks=false]{hyperref}

\iccvfinalcopy 

\ificcvfinal\fi

\begin{document}

\setlength{\abovedisplayskip}{3pt}
\setlength{\belowdisplayskip}{3pt}

\makeatletter
\newcommand{\thickhline}{%
    \noalign {\ifnum 0=`}\fi \hrule height 1pt
    \futurelet \reserved@a \@xhline
}

\makeatother
\newcommand{\MATTHIAS}[1]{{\emph{\textcolor{red}{\textbf{Matthias:~#1}}}}}
\newcommand{\ANGIE}[1]{{\emph{\textcolor{blue}{Angie: #1}}}}
\newcommand{\JUSTUS}[1]{{\emph{\textcolor{magenta}{Justus: #1}}}}
\newcommand{\YAWAR}[1]{{\emph{\textcolor{ForestGreen}{Yawar:~#1}}}}
\newcommand{\FANGCHANG}[1]{{\emph{\textcolor{brown}{Fangchang:~#1}}}}
\newcommand{\TODO}[1]{{\emph{\textcolor{BrickRed}{TODO: #1}}}}

\title{RetrievalFuse: Neural 3D Scene Reconstruction with a Database}

\author{
Yawar Siddiqui$^1$~~~
Justus Thies$^{1,2}$~~~
Fangchang Ma$^3$~~~
Qi Shan$^3$~~~
Matthias Nie{\ss}ner$^1$
Angela Dai$^1$~~~
\vspace{0.2cm} \\ 
$^1$Technical University of Munich~~~
$^2$Max Planck Institute for Intelligent Systems, Tübingen~~~
$^3$Apple
\vspace{0.2cm} \\ 
}

\twocolumn[{%
	\renewcommand\twocolumn[1][]{#1}%
	\maketitle
	\begin{center}
	    \vspace{-0.65cm}
		\includegraphics[width=\linewidth]{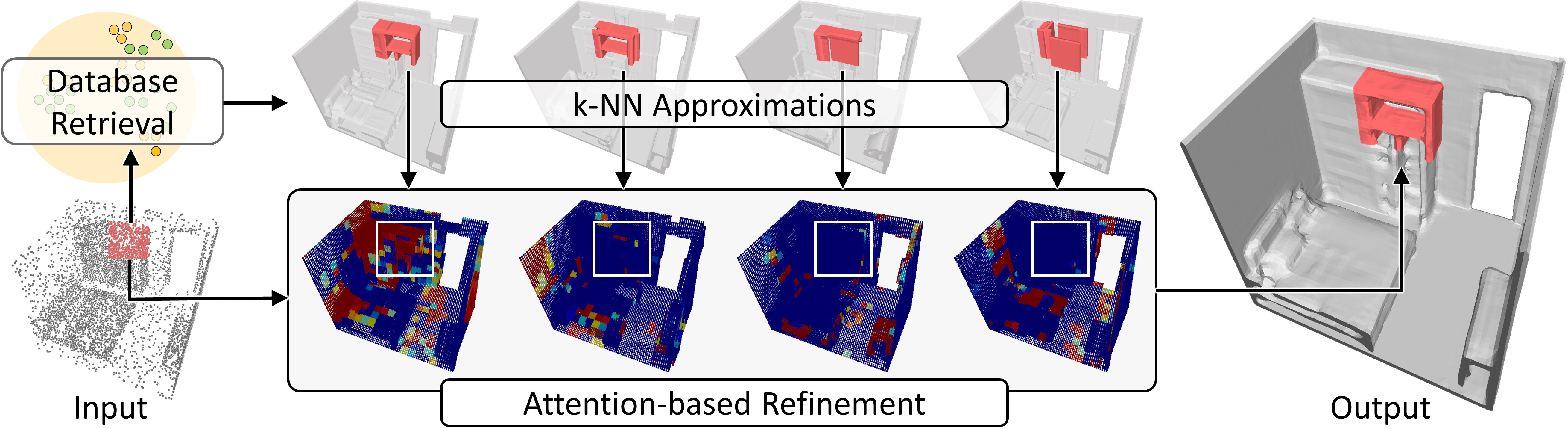}
	    \vspace{-0.515cm}
		\captionof{figure}{
		We present a new approach for 3D reconstruction conditioned on sparse point clouds or low-resolution geometry. 
		Rather than encoding the full generative process in the neural network, which can struggle to represent local detail, we leverage an additional database of volumetric chunks from train scene data. 
		For a given input, multiple approximate reconstructions are first created with retrieved database chunks, which are then fused together with an attention-based blending~-- facilitating transfer of coherent structures and local detail from the retrieved train chunks to the output reconstruction.
		}
		\label{fig:teaser}
	\end{center}    
	\vspace{0.15cm}
}]

\maketitle
\ificcvfinal\thispagestyle{empty}\fi


\begin{abstract}
3D reconstruction of large scenes is a challenging problem due to the high-complexity nature of the solution space, in particular for generative neural networks.
In contrast to traditional generative learned models which encode the full generative process into a neural network and can struggle with maintaining local details at the scene level, we introduce a new method that directly leverages scene geometry from the training database.
First, we learn to synthesize an initial estimate for a 3D scene, constructed by retrieving a top-$k$ set of volumetric chunks from the scene database. 
These candidates are then refined to a final scene generation with an attention-based refinement that can effectively select the most consistent set of geometry from the candidates and combine them together to create an output scene, facilitating transfer of coherent structures and local detail from train scene geometry.
We demonstrate our neural scene reconstruction with a database for the tasks of 3D super resolution and surface reconstruction from sparse point clouds, showing that our approach enables generation of more coherent, accurate 3D scenes, improving on average by over 8\% in IoU over state-of-the-art scene reconstruction.
\end{abstract}

\section{Introduction}
3D scene reconstruction has been a long-standing problem in computer vision and graphics, and has recently seen a renewed flurry of developments, driven by successes in generative neural networks \cite{deepsdf,mescheder2019occupancy,dai2020sg,peng2020convolutional}.
In particular, developing an effective geometric reconstruction is challenging due to the dimensionality of the problem, and the simultaneous expressibility for local details as well as coherent, complex global structures.
In recent years, various approaches have been developed for geometric reconstruction, encoding the full generative process into a neural network. This can result in difficulty in representing large-scale, complex scenes, as all levels of detail must be fully encoded as part of the generative network. 

We thus propose to augment geometric reconstruction with a database which our method learns to leverage at inference time, and introduce a generative model that does not need to encode the entire training data as part of the network parameters. Instead, our model learns how to best transfer structures and details from retrieved scene database geometry.

We construct this database as geometric, cropped chunks of 3D scenes from train scene data.
Each chunk represents clean, consistent, high-resolution geometry.
We leverage these chunks as a basis for scene reconstruction.

To this end, we develop a neural 3D scene reconstruction approach to generate 3D scenes as volumetric distance fields. This approach consists of two main steps: a top-k nearest neighbor retrieval and combination for initial estimation, and a refinement stage to produce the higher-quality, final reconstruction.
Specifically, to generate a 3D scene from an input condition (e.g., a noisy or sparse observation of a scene), we first learn to construct an initial estimate of the scene as a combination of cropped volumetric chunks from the database.
By providing an initial estimate based on chunks of existing scene geometry, we can more easily encourage consistent, sharp structures already seen in the existing scene geometry.
Since these initial scene crop estimates may not be entirely locally consistent with each other, we then refine this estimate to produce a final scene reconstruction.
The scene refinement is based on patch-based attention which encourages the selection of given scene chunk estimates where they suffice -- maintaining their clean details -- and synthesizing refined geometry otherwise.

By leveraging database retrieval in combination with a generative model, our approach does not need to encode the full train set for effective reconstruction, and facilitates generation of globally coherent, high quality 3D scenes.
We demonstrate our approach on the tasks of 3D super resolution and 3D surface reconstruction from sparse point samples on both synthetic and real-world 3D scene data, showing significant qualitative and quantitative improvement in comparison to state-of-the-art reconstruction approaches.
Additionally, we show that our approach can also be applied to other generative representations, in particular, to improve implicit-based reconstruction.

In summary, our main contributions are:
\begin{itemize}
    \item A neural 3D reconstruction technique that leverages details present in a database of cropped scene chunks for improving reconstructed geometry.
    \item A patch-wise attention-based refinement that robustly fuse together details from the retrieved scene chunks.
\end{itemize}

\section{Related Works}

\paragraph{Learned 3D Shape Reconstruction.} 
3D shape reconstruction is a long-standing problem in computer vision. 
We refer readers to Szeliski~\cite{szeliski2010computer} for a more comprehensive review of the classic techniques. 
Recently, inspired by the progress of deep learning for images, many developments have been made in deep generative models for reconstructing 3D shapes, largely focusing on leveraging different geometric representations.

Early generative neural networks focused on voxel grids as a natural extension of pixels, with a regular structure well-suited for convolutions, but can struggle with cubic growth in dimension \cite{maturana2015voxnet,wu20153d,choy20163d,dai2017shape}.
Multi-resolution representations were proposed~\cite{hane2017hierarchical, tatarchenko2017octree} to address the cubic complexity with hierarchical data structures.
Rather than operating on a regular grid, point cloud based approaches propose to generate points only on the geometric surface \cite{fan2017point,yang2019pointflow}, but do not encode structural connectivity. 
Mesh-based approaches have also been proposed to efficiently capture surface geometry while encoding connectivity, but tend to rely on strong topological assumptions such as a template mesh that is then deformed~\cite{wang2018pixel2mesh}, or a small number of vertices for free-form generation~\cite{dai2019scan2mesh}.
Implicit representations encoded directly by the neural network enable modeling of a continuous surface, typically as binary occupancies or signed distance fields \cite{mescheder2019occupancy,deepsdf,chibane2020implicit}; such representations have seen notable success in modeling single objects but can struggle to directly scale to scenes.

\paragraph{Learned 3D Scene Reconstruction.} 
Compared to shape reconstruction, scene-level reconstruction is significantly more challenging due to the scale, variance, and complexity of geometry. 
Several approaches have been proposed to combine local implicit functions with a coarse volumetric basis \cite{jiang2020local,deep_local_shapes,peng2020convolutional} to capture complex, large-scale scene reconstructions.
SG-NN~\cite{dai2020sg} leverages a single, sparse volumetric network for large-scale scene completion in a self-supervised fashion. These approaches rely on encoding the full generative process into network parameters, whereas we leverage a basis of existing scene geometry, that does not need to be fully encoded but rather refined to transfer desired geometric characteristics from the valid scene geometry (e.g., clean structures, local details). 

\paragraph{2D/3D Retrieval.} Our approach is related to 2D image retrieval and completion applications~\cite{datta2008image}, where recent work~\cite{radenovic2018fine, xu2020texture} focuses on developing a CNN to automatically retrieve relevant patches from a large collection of unordered images. 
Note that memorization is also an active area of research in language models~\cite{khandelwal2019generalization}.

For 3D retrieval, the pioneering work of Chen \etal~\cite{chen2003visual} proposed a 3D shape retrieval system based on visual similarity.
More recently, several works have been proposed to leverage 3D CAD model retrieval to represent objects in input images or 3D scans \cite{li2015database,izadinia2017im2cad,avetisyan2019scan2cad,kuo2020mask2cad,izadinia2020licp}, but are limited to the objects in the CAD dataset, while we use our retrieval as a basis for enabling more accurate reconstruction from learned selection and blending of retrieved scene geometry.

\section{Method}
We formulate the problem as a general 3D reconstruction conditioned on inputs which can be spatially correlated with the output scene. This can be instantiated into applications like 3D super-resolution from low-resolution observations and surface reconstruction from a sparse set of 3D point measurements. 
The proposed approach augments a generative model for synthesizing 3D scenes with external knowledge in the form of a database of existing 3D scene data. 
Specifically, a typical learning-based 3D reconstruction function, $f_r$, trains on pairs $\{\mathbf{x}_i, \mathbf{y}_i\}$ of input and ground truth 3D data, with a loss to measure the distance between each $f_r(\mathbf{x}_i)$ and $\mathbf{y}_i$. 
At inference time, given an unseen input $\mathbf{x}_j$, $f_r$ is applied as $f_r(\mathbf{x}_j) = \hat{\mathbf{y}}_j$, without using any additional information. 

In contrast, our approach maintains the training data $\{\mathbf{y}_i\}$ to form a basis of an initial reconstruction estimate during inference. 
An overview of our approach is visualized in Fig.~\ref{fig:teaser}.
We first learn to retrieve similar train data to the input condition to construct multiple initial reconstruction estimates, $\mathbf{x}_j\rightarrow \{\mathbf{y}_j'\}$. 
We then refine these estimates to produce the final reconstruction, $\mathbf{x}_j,\{\mathbf{y}_j'\} \rightarrow \mathbf{y}_j$. 
This facilitates transfer of scene geometry characteristics such as detail and global structures from train data to produce more coherent and detailed output reconstructions.

We demonstrate our approach on the 3D reconstruction tasks of super resolution and surface reconstruction from points, learning to reconstruct a distance field representation of an output 3D scene from a low-resolution distance field and point cloud, respectively. 
For a set of train scenes $\{\mathbf{y} \in \mathbb{R}^{L \times W \times H}\}$, we consider cropped scene chunks $\{y_i \in \mathbb{R}^{l \times w \times h}\}$ as our additional knowledge database. 
We first learn to map spatially corresponding input chunks $\{x_i\}$ to these $\{y_i\}$ by constructing a shared embedding space between the $x_i$ and $y_i$ and retrieving the $k$ nearest neighbors for a new $x_j$. These nearest neighbors then form candidates for a scene reconstruction.

Based on the input condition and these candidates, which comprise $k$ distance fields for each chunk in the output scene, we then learn to refine the initial estimates to a final distance field scene reconstruction.
The initial basis constructed by existing 3D scene data is composed of chunks of valid local and global structures (e.g., flat walls or floors, full structures as well as sharp details of objects), enabling our refinement to more easily maintain these characteristics in the output reconstruction.
To encourage the transfer of desired global structures and fine details to the final reconstruction, we employ attention to help select the most meaningful parts of the initial estimate.
This facilitates coherent reconstructions while maintaining local detail.

\subsection{Estimating Reconstruction as a Composition of Existing Scene Data}

\begin{figure}
	\centering
	\includegraphics[width=\linewidth]{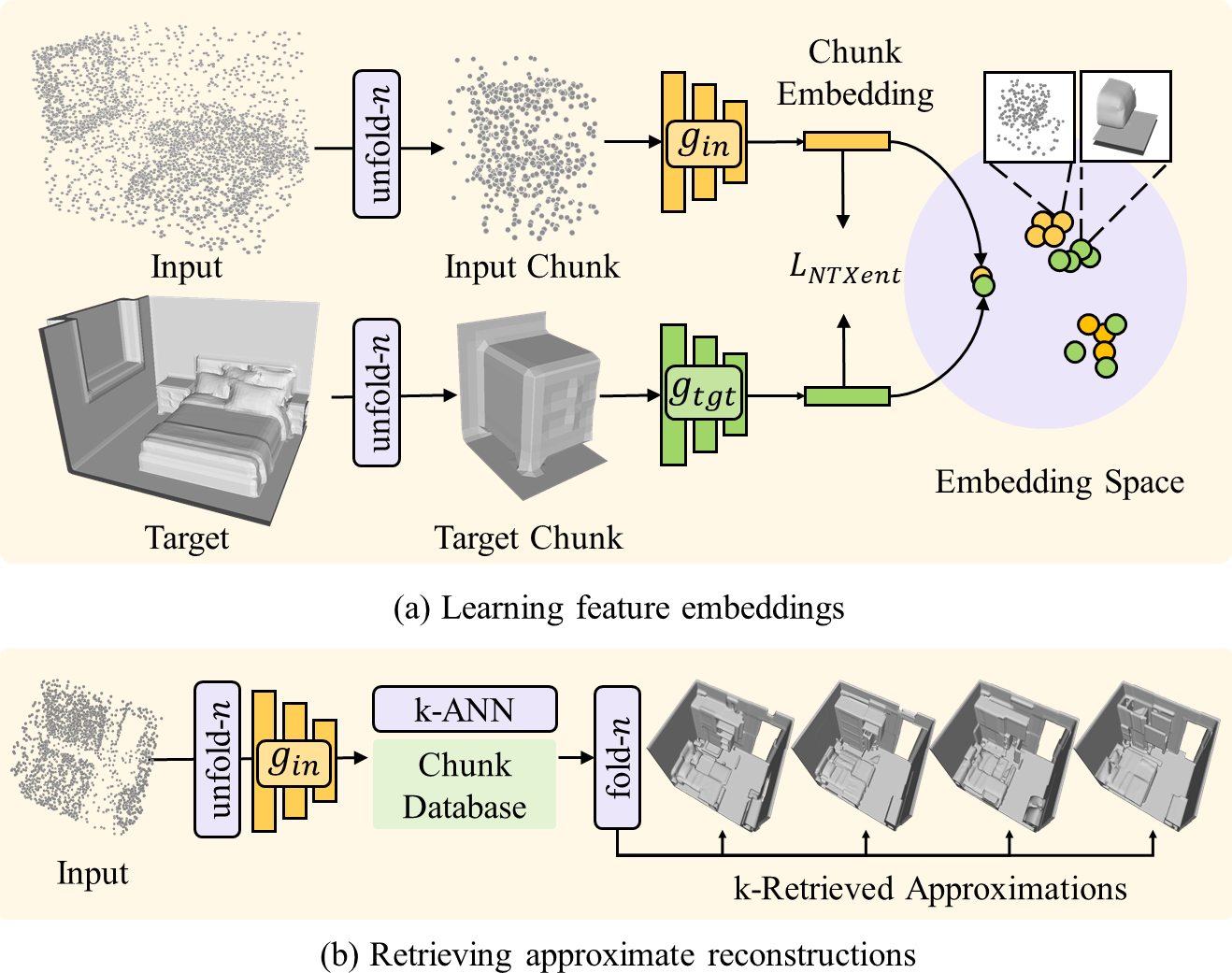}
	\vspace{-0.6cm}
	\caption{
	Estimating reconstruction with database retrievals. 
	(a) Input and target scenes are decomposed into a total of $n^3$ chunks each by \textit{unfold-n}; input/target chunks are embedded into a shared space which is trained using a contrastive loss.
	The database is composed of embedded target chunks from the train set, and used for retrieval for new input queries.
	(b) For a new input, the $k$-NN retrieved chunks create approximate reconstructions, which can then be refined.
	}
	\label{fig:method_retrieval}
	\vspace{-0.25cm}
\end{figure}

We first aim to approximate a scene reconstruction as a composition of cropped chunks of existing scene data from the database $\{y_i\}$.
By recomposing cropped chunks of scene data, we can express diverse scene content while leveraging a basis of existing scene data.
To construct this approximation, we learn to retrieve $k$ candidate chunks from the database, providing a variety of candidate reconstruction estimates that can be used to inform the final reconstruction refinement.
These multiple candidates provide alternatives to the following refinement stage, as we cannot expect to have exactly corresponding chunk geometry at test time.
We show an overview of our retrieval-based reconstruction estimation in Fig.~\ref{fig:method_retrieval}.

To find the best candidate chunks from the database for the corresponding part of the input, we learn to embed chunks of input observations $\{x_i\}$ and target scene chunks $\{y_i\}$ into a shared latent space, where top-$k$ nearest neighbor retrieval is then performed.
We thus embed $x_i$ into a 64 dimensional latent space by passing it through a stack of convolutional layers followed by a fully connected layer, resulting in feature $g_{x_i} = g_{in}(x_i)$. We similarly embed the corresponding target $y_i$ into a 64 dimensional latent space by also passing it through a stack of convolutional layers with a fully connected layer at the end, resulting in feature $g_{y_i} = g_{tgt}(y_i)$.
Inspired by contrastive learning~\cite{hadsell2006dimensionality}, we construct the shared space using a normalized, temperature-scaled cross entropy loss (NTXent)~\cite{chen2020simple}:
\begin{equation}\small
    L^\text{NTXent} = -\log\frac{\exp\left(g_{x_i}\!\cdot\! g_{y_i}/\tau\right)}{\sum^{N}_{k=1}\mathbbm{1}_{[k\neq{i}]}\exp\left(g_{x_i}\!\cdot\!g_{y_i}/\tau'(\tau,y_i, y_k)\right)}
\end{equation}
where $N$ denotes the number of samples in the minibatch, $\mathbbm{1}_{[k\neq{i}]}$ evaluating to 1 iff $k\neq i$, and $\tau \in (0, 1]$ is a temperature parameter.
This encourages similarly structured target scene chunks to be retrieved for an input observation.

A minibatch may contain target chunk $y_k$ similar in geometry to the target chunk $y_i$ where $k\neq i$. 
We thus use $\tau'$ to discourage  heavy penalization in this scenario, by making the temperature scaling to be a function of IoU between the target chunks,
\begin{equation}
    \tau'(\tau,y_i, y_k) = \tau + (1 - \tau)\,\sigma\!\left(a\!\cdot\!\mathrm{IoU}\left(y_i, y_k\right) + b\right)
\end{equation}
where $a$ and $b$ are constant shift and bias, and $\sigma$ a sigmoid. 

\paragraph{Retrieval Database.} 

Once the networks have been trained, the target chunks $\{y_i\}$ are all embedded as $g_{y_i}$ into the latent space to support chunk retrieval.
Then for a new input observation $\mathbf{x}$, it is split into spatial chunks $\{x_j\}$, for which $k$ nearest neighbors are found from $g_{x_j}$ by an $\ell_2$ distance metric.
This provides $k$ candidate reconstruction estimates $\{\mathbf{y}'\}$.

\subsection{Reconstruction Refinement}
\begin{figure}
	\centering
	\includegraphics[width=\linewidth]{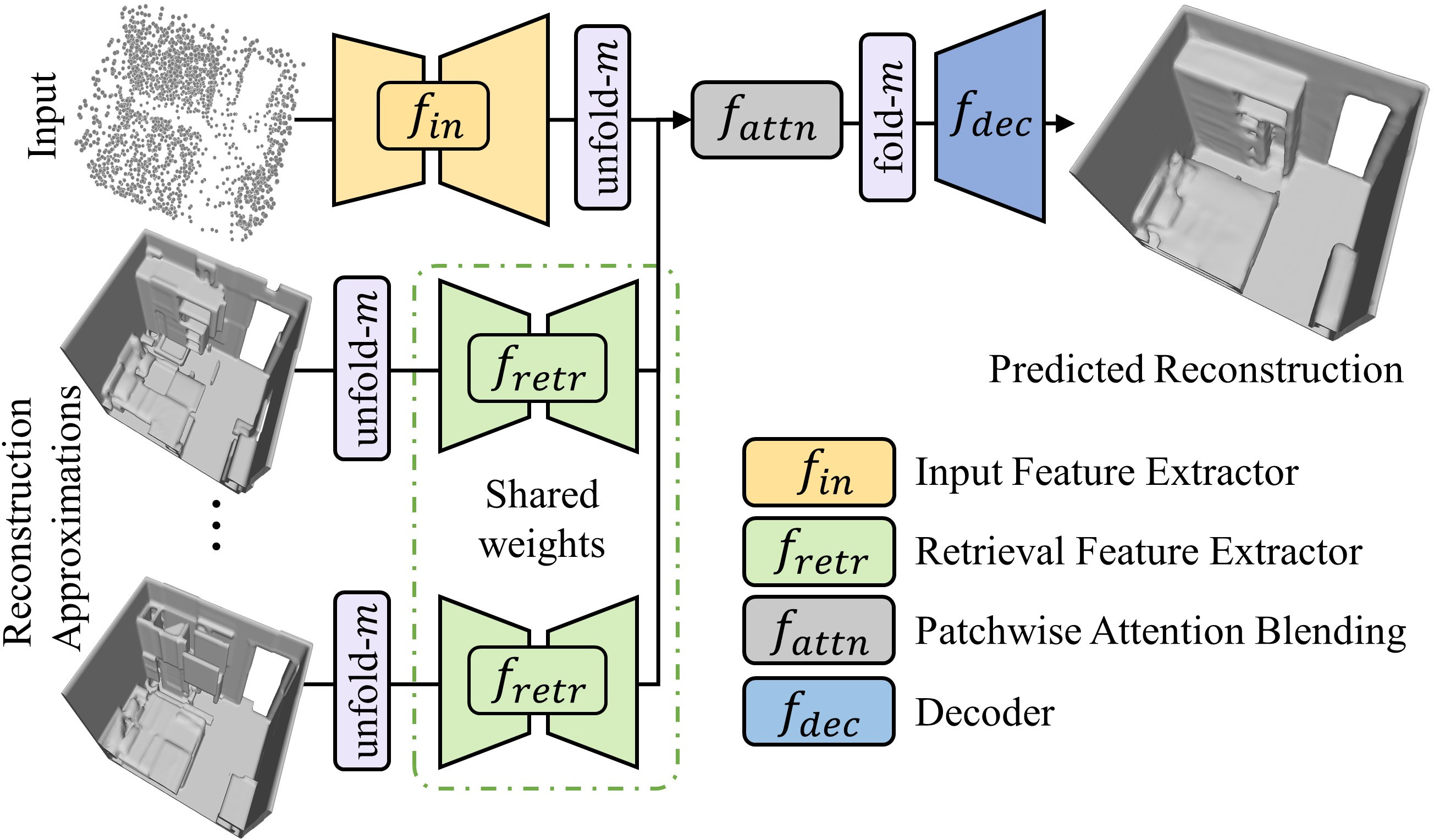}
	\vspace{-0.7cm}
	\caption{
	The input and reconstruction approximations are passed through feature extractors. 
	The resulting input feature grid is split into patches spatially aligned with the patch features from the retrieval approximations, which are then fused together with our attention blending network. 
	Finally, the patch-wise blended features are re-interpreted as a full feature volume and decoded to output geometry.
	}
	\label{fig:method_refinement}
	\vspace{-0.25cm}
\end{figure}
Our initial retrieval-based reconstruction estimate provides a strong prior for  global structures and fine-scale details in the scene, but the retrieved chunks may not be fully locally consistent with each other. Therefore, we leverage this estimated reconstruction to refine a globally coherent reconstruction while maintaining local detail.
We visualize this refinement in Fig.~\ref{fig:method_refinement}.
The input observation $\mathbf{x}$ and the estimated retrieval-based reconstructions $\{\mathbf{y}'\}$ inform the final refinement. 
The input is passed through a U-Net~\cite{ronneberger2015u}-based feature extractor $f_{in}$ to produce a grid of features, which is split into a set of patch features. The retrieval approximations are first split into volumetric patches and are passed through feature extractor $f_{retr}$, analogously structured as $f_{in}$ but smaller in parameters since it operates on retrieved patches.
Next, we blend together the features from the input with those from corresponding retrieval approximations.
We leverage a patch-based attention layer which learns to select and blend the retrieved patches, based on feature similarities to the spatially-corresponding features from the input data.
The resulting feature grid is finally decoded with convolutions to output the reconstructed geometry. 

\paragraph{Patch-based Attention.} We leverage a patch-based attention to encourage selection of robust patches from the retrieved chunks to inform the final reconstruction, i.e., only features that would most help the reconstruction are used.
\begin{figure}
	\centering
	\includegraphics[width=\linewidth]{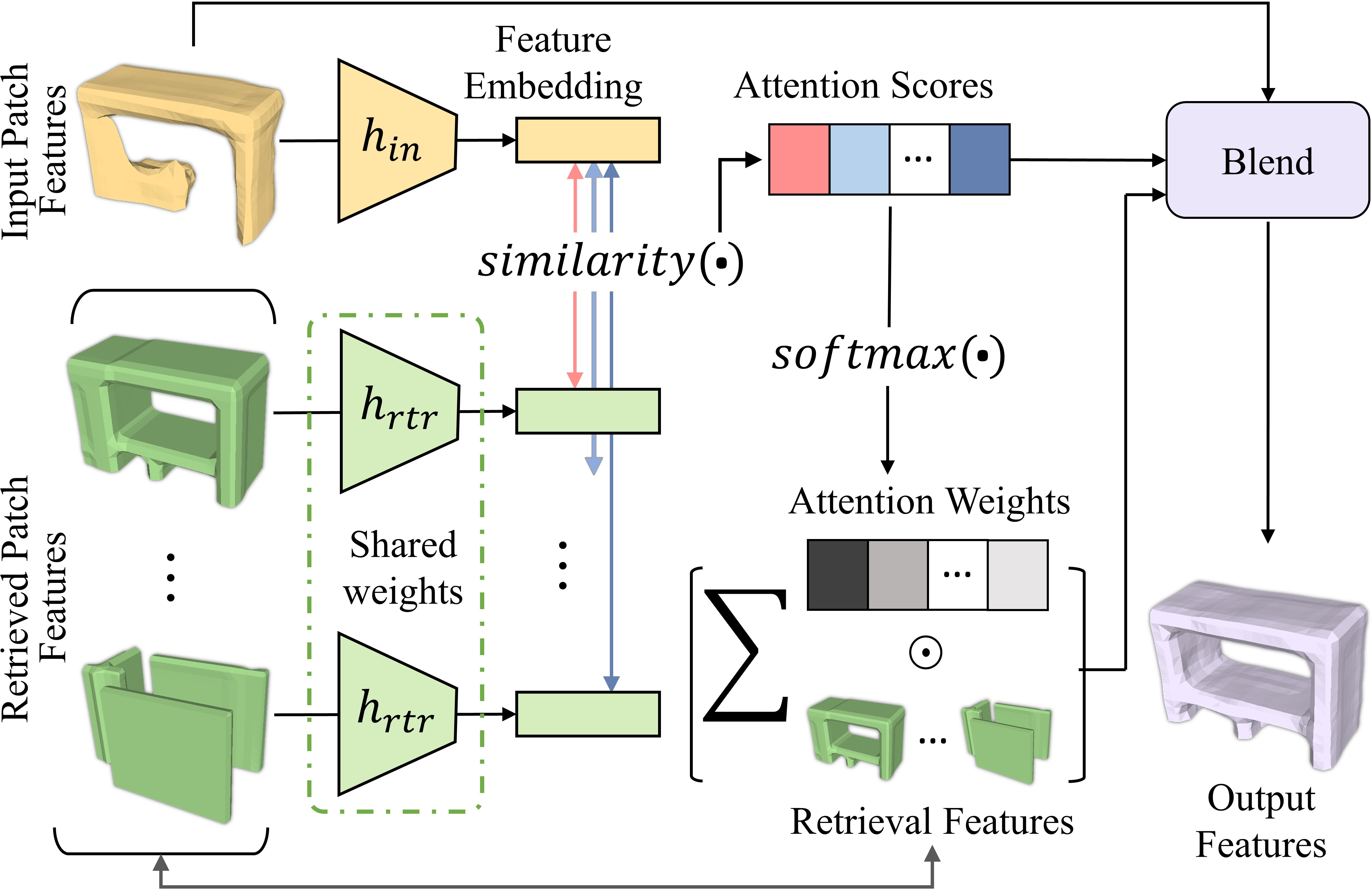}
	\vspace{-0.5cm}
	\caption{Feature similarity between input and retrieved patch features informs attention scores. Attention weights derived from the scores determine the contribution among the retrievals. A learned blending function then fuses input and retrieval features based on the max attention score.}
	\label{fig:method_attention}
	\vspace{-0.25cm}
\end{figure}
The features from the input $\mathbf{x}$ and from the retrieval-based reconstructions $\{\mathbf{y}'\}$ can be spatially aligned with each other.
To select the most relevant features, we consider patches of these aligned features (with the patch size smaller than the chunk size of the retrieval, as we want to be able to select features within retrieved chunks); each patch of the input then corresponds to $k$ patches of retrieved chunks.
Similarity between input and retrieval patch features is then computed in a lower-dimensional projection space, using cosine similarity to compute attention scores. 
%
The process is visualized in Fig.~\ref{fig:method_attention}.

More formally, if $p_{in}$ and $p_{retr_i}$ are the input and retrieved patch features for the $i^{th}$ nearest neighbor retrieval, the patched attention layer first computes the attention score as
\begin{equation}
    s_i = s(p_{in}, p_{retr_i}) = h_{in}(p_{in}) \cdot h_{retr}(p_{retr_i})
\end{equation} where $h_{in}$ and $h_{retr}$ are networks implemented as MLPs that project $p_{in}$ and $p_{retr_i}$ to a 32-dimensional normalized space.
The attention weights are computed as softmax of attention scores:
\begin{equation}
    w_i = \frac{\exp(Cs_i)}{\sum_{j=1}^{k}\exp(Cs_j)}
\end{equation} where $C$ is a hyperparameter controlling sharpness of the softmax.
$C$ encourages selection over blending from the $k$ retrievals, in order to maintain the local detail present in the retrieved patches.
The total contribution due to the $k$ retrievals is then given by the attention weighted sum of retrieval features.
Next, a learned blending function blends the input patch feature with this weighted sum based on the maximum attention score.
That is, once we have the attention weights, the output from attention layer is given as 
\begin{equation}
    (1 - \beta)\,p_{in} + \beta\sum_{i=1}^{k}w_i\,p_{retr_i}
\end{equation} with the blending coefficient given as 
\begin{equation}
    \beta = \beta(s_1, s_2, ..., s_k) = \mathrm{sigmoid}(c\cdot\max_i s_i + d),
\end{equation} $c$ and $d$ are learnable shift and bias parameters. Intuitively, the attention weights determine which of the retrievals should contribute, while $\beta$ determines how much input features should contribute compared to retrieval features.
Finally, the blended patches are reinterpreted as a full grid and decoded to an output distance field.

\paragraph{Refinement Loss.}
To train the refinement, we employ a reconstruction loss on the final prediction as well as a retrieval-reconstruction loss and attention loss:
\begin{equation}
    L_{\mathrm{ref}} = L_{\mathrm{recon}} + \lambda_{\mathrm{retr}}\,L_{\mathrm{retr}} + \lambda_{\mathrm{attn}}\,L_{\mathrm{attn}}.
\end{equation}
$L_{\mathrm{recon}}$ denotes the reconstruction loss on the final predicted distance field $\mathbf{y}_{\textrm{recon}}$ with the ground truth distance field $\mathbf{y}_{\textrm{gt}}$ as an $\ell_1$ loss:
\begin{equation}
    L_{\mathrm{recon}} = |\mathbf{y}_{\textrm{recon}} - \mathbf{y}_{\textrm{gt}}|_1.
\end{equation} 
$L_{\mathrm{retr}}$ ensures that the refinement decoder continues to decode to the original distance field of the retrieved chunk features: 
\begin{equation}
    L_{\mathrm{retr}} = |f_{dec}(f_{retr}(y_j)) - y_j|_1
\end{equation}
where $y_j$ is a chunk of $\mathbf{y}_{\textrm{gt}}$.
Finally, attention embedding space is supervised with
\begin{equation}
    L_{\mathrm{attn}} = \mathrm{NTXent}(h_{in}(p_{x_j}), h_{retr}(p_{y_j})) 
\end{equation} 
where NTXent is the normalized cross entropy loss, and $p_{y_j}$, $p_{x_j}$ are target and corresponding input patch features respectively. 

\subsection{Implementation Details}
For both tasks of 3D super-resolution and point cloud to surface reconstruction, we use a $64^3$ truncated distance field (TDF) representation for the target geometry (larger scenes are processed in a sliding window fashion in $64^3$ windows), which are converted to meshes by Marching Cubes~\cite{lorensen1987marching}. 
We use a $16^3$ chunk size in the target domain for retrievals, resulting in $4\times4\times4=64$ chunks per sample. 
The spatial attention uses a smaller patch size of $4^3$. 

We use a temperature of $0.2$ for the retrieval NTXent, and $0.05$ for the attention phase NTXent. We found that a lower temprature works better for the smaller sized patches.
The refinement phase uses $k = 4$ retrieval approximations for an input.
The refinement loss coefficients $\lambda_{\mathrm{retr}} = 0.5$ and $\lambda_{\mathrm{attn}} = 0.05$ to bring the losses to similar magnitudes.

All networks are trained using Adam~\cite{kingma2014adam} with a learning rate of $10^{-4}$. We use a batch size of 196 for retrieval training, and 8 for refinement. 
We train on a single NVIDIA 2080Ti for 150k iterations for retrieval ($\approx 10$ hours) and 350k iterations for refinement ($\approx 40$ hours)

\section{Results}
We demonstrate our approach on the tasks of 3D super-resolution and surface reconstruction from sparse point clouds, on both objects and scenes as well as synthetic and real-world data.
We evaluate on three datasets with increasing complexity: ShapeNet~\cite{chang2015shapenet} (synthetic shapes), 3DFront~\cite{fu20203dfront} (synthetic scenes) and Matterport3D~\cite{chang2017matterport3d} (real-world 3D scans).
For ShapeNet, we use the 13 class subset and train/test split from \cite{choy20163d}. 
For 3DFront, we use the scenes which have furniture in a train/test split of $15000/2850$ rooms.
We use the official train/test split for Matterport3D of 72/18 buildings and 1799/394 rooms.

\smallskip
\noindent \textbf{Metrics.}
We evaluate the reconstructed geometry with $4$ complementary metrics: Volumetric IoU (IoU), Chamfer $\ell_1$ Distance (CD) in meters, Normal Consistency (NC) as cosine distances and F-score (F1) at $1\%$ window size threshold. 
Additional evaluation details can be found in the appendix.

\subsection{3D Super Resolution}
For the task of 3D super resolution, we consider low-resolution geometry as input, and aim to reconstruct high-resolution target geometry.
We use input / target voxel sizes of $0.434$m / $0.054$m for 3DFront, $0.15$m / $0.0375$m for Matterport3D, and resolutions $8^3$ / $64^3$ for ShapeNet.
For synthetic and real-world 3D scenes of varying sizes, we operate in a sliding window fashion with a stride of $64$ voxels; we similarly run all baselines in the same sliding fashion.

\begin{figure*}
	\centering
	\includegraphics[width=\linewidth]{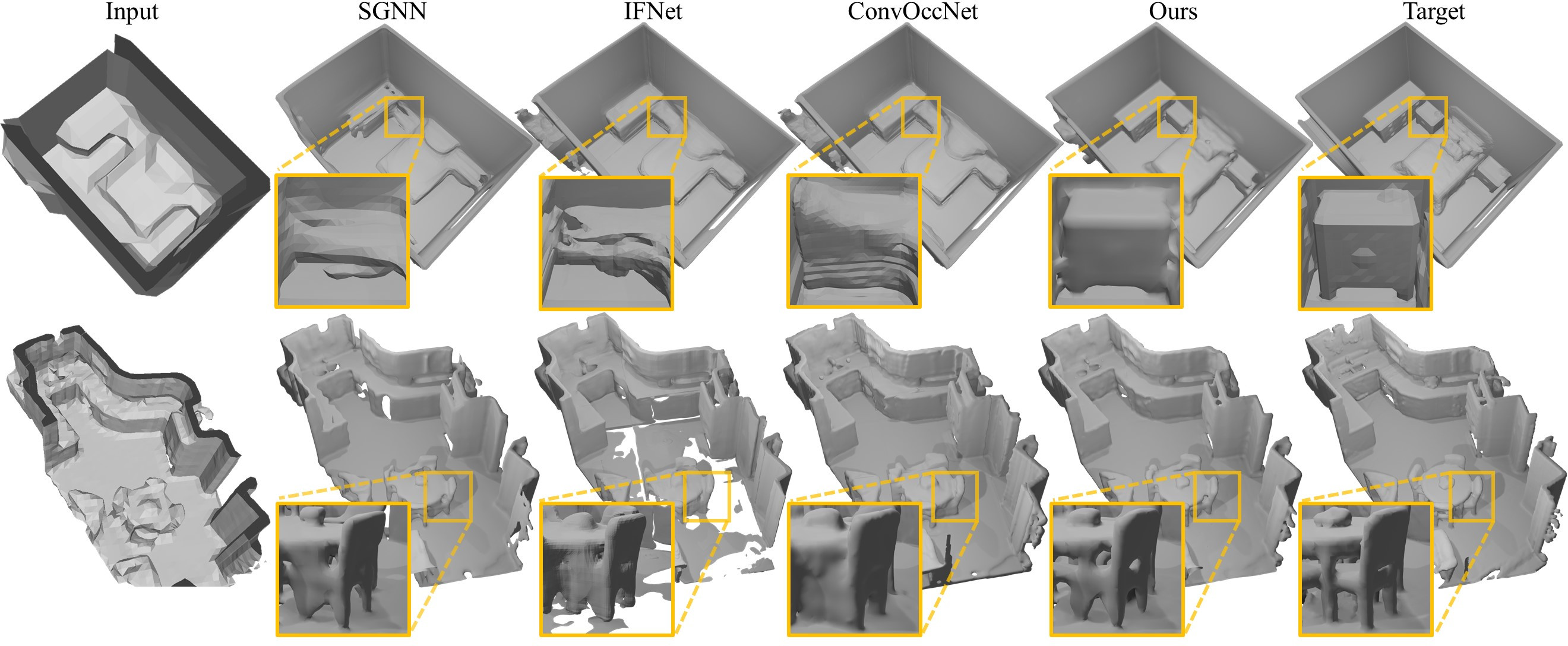}
    \vspace{-0.6cm}
	\caption{
	    3D super resolution on 3DFront (top) and Matterport3D (bottom) datasets. In contrast to other approaches, our method generates more coherent 3D geometry with sharper details.
    }
	\label{fig:experiments_superresolution}
	\vspace{-0.25cm}
\end{figure*}

\smallskip
\noindent \textbf{Comparison to state of the art.}
We compare to state-of-the-art 3D generative approaches: SG-NN~\cite{dai2020sg} which operates on sparse volumetric data, IFNet~\cite{chibane2020implicit} which learns implicit reconstruction, and Convolutional Occupancy Networks~\cite{peng2020convolutional} which uses a hybrid of volumetric convolutions coupled with implicit decoders. While these methods encode the entire generative process in the network, we additionally use an explicit database that can assist the reconstruction during inference.
In Tab.~\ref{tab:superres}, we show a quantitative comparison; our ability to leverage strong priors from retrieved scene data enables more effective reconstructions of 3D scenes.
We additionally show a qualitative comparison in Fig.~\ref{fig:experiments_superresolution}; our approach maintains sharper detail which is more easily propagated by retrieving priors from the train set.
{
\begin{table*}[tp]
\begin{center}
\small
\begin{tabular}{|l|l|c|l|l|l|l|l|l|l|l|l|l|}  
\hline
\multirow{2}{*}{Method} & \multicolumn{4}{c|}{ShapeNet} & \multicolumn{4}{c|}{3DFront} & \multicolumn{4}{c|}{Matterport3D}  \\ 
\cline{2-13}
                        & IoU$\uparrow$ & CD$\times 10^{-2}$$\downarrow$ & F1$\uparrow$ & NC$\uparrow$            & IoU$\uparrow$ & CD$\downarrow$ & F1$\uparrow$ & NC $\uparrow$           & IoU$\uparrow$ & CD$\downarrow$ & F1$\uparrow$ & NC$\uparrow$ \\
\hline
SGNN                    & 0.624 & 0.668 & 0.813 & 0.889         & 0.639 & 0.032 & 0.733 & 0.900              & 0.731 & 0.021 & 0.697 & 0.916                \\ 
ConvOcc                 & 0.648 & 0.726 & 0.838 & \textbf{0.906}            & 0.631 & 0.033 & 0.711 & 0.901             & 0.584 & 0.027 & 0.542 & 0.879                    \\ 
IFNet                   & 0.650 & 0.623 & 0.838 & 0.892              & 0.639 & 0.041 & 0.736 & 0.878             & 0.593 & 0.028 & 0.624 & 0.893              \\ 
\hline
Ours                    & \textbf{0.655} & \textbf{0.590} & \textbf{0.844} & 0.905               & \textbf{0.751} & \textbf{0.027} &\textbf{0.801} & \textbf{0.922}              &  \textbf{0.739} & \textbf{0.020} & \textbf{0.708} & \textbf{0.923}                  \\
\hline
\end{tabular}
\caption{Evaluation of reconstruction performance on 3D super-resolution on ShapeNet, 3DFront, and Matterport3D, with $8\times$ higher target resolution for synthetic data and $4\times$ higher resolution for real data.}
\label{tab:superres}
\end{center}
\vspace{-0.15cm}
\end{table*}
}

\subsection{Surface Reconstruction}
We additionally demonstrate our approach on the task of surface reconstruction from point cloud data.
Input point clouds are obtained by randomly sampling $500$ points for each shape in ShapeNet, $1000$ points per $3.45^3\mathrm{m}^3$ for 3DFront, and $1000$ points per $2.4^3\mathrm{m}^3$ for Matterport3D.

\begin{figure*}
	\centering
	\includegraphics[width=\linewidth]{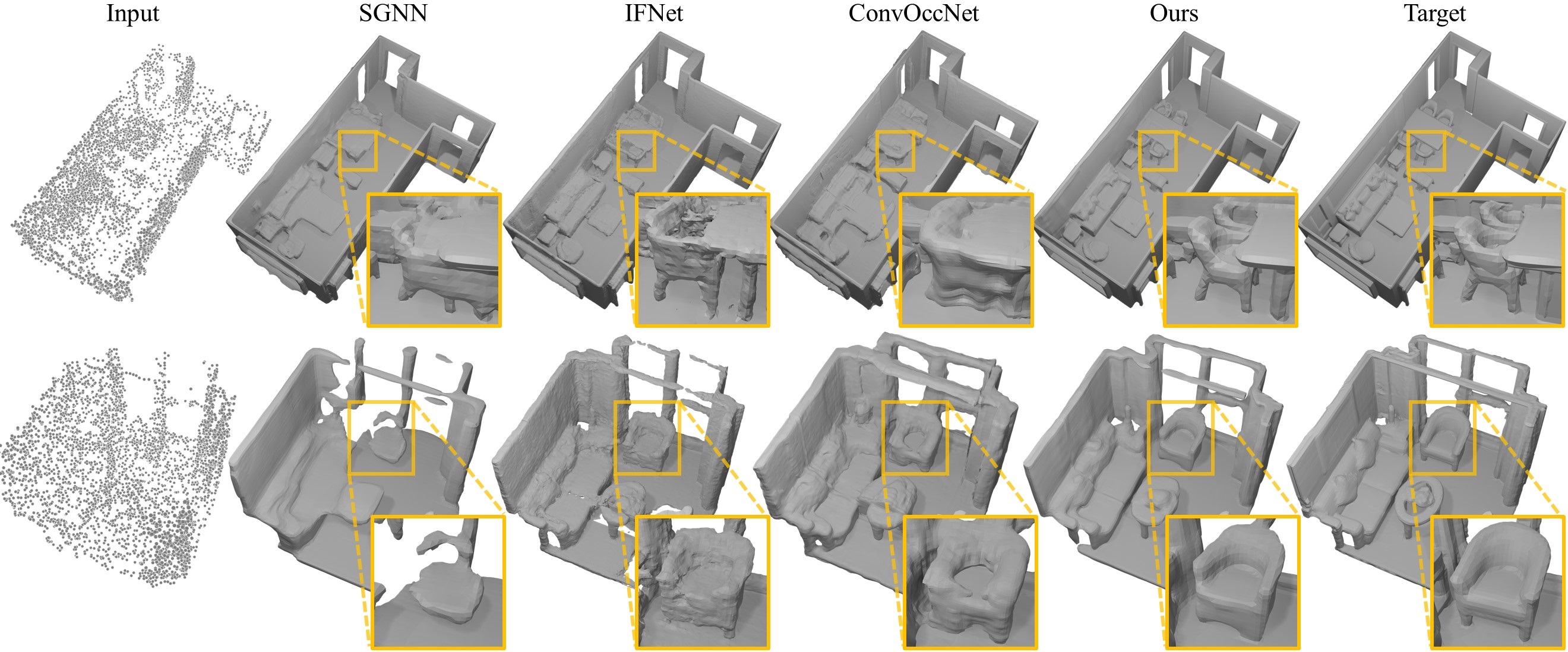}
	\vspace{-0.6cm}
	\caption{
	    Point cloud to surface reconstruction on 3DFront (top) and Matterport3D (bottom) datasets.
	    Our approach captures more coherent structures and object details.
    }
	\label{fig:experiments_surface_reconstruction}
	\vspace{-0.25cm}
\end{figure*}

\smallskip
\noindent \textbf{Comparison to state of the art.}
We compare to the state-of-the-art 3D generative approaches as in the 3D super-resolution task (SG-NN~\cite{dai2020sg}, IFNet~\cite{chibane2020implicit}, Convolutional Occupancy Networks~\cite{peng2020convolutional}), in addition to Screened Poisson Surface Reconstruction (SPSR)~\cite{kazhdan2006poisson,kazhdan2013screened} and Local Implicit Grids (LIG)~\cite{jiang2020local}.
All data-driven methods are trained on our data.
For our method, SG-NN, and IFNet which take volumetric input, we consider the point cloud as a volumetric occupancy grid (occupancy for voxels containing any points). 
Tab.~\ref{tab:pcrecon} shows a quantitative comparison, where our learned use of train scene data through an attention-based refinement provides more accurate geometric reconstruction.
Fig.~\ref{fig:experiments_surface_reconstruction} additionally shows that our reconstructions more effectively capture both global structures and local details in the scenes.
{
\begin{table*}[tp]
\begin{center}
\small
\begin{tabular}{|l|l|c|l|l|l|l|l|l|l|l|l|l|} 
\hline
\multirow{2}{*}{Method} & \multicolumn{4}{c|}{ShapeNet} & \multicolumn{4}{c|}{3DFront} & \multicolumn{4}{c|}{Matterport3D}  \\ 
\cline{2-13}
                        & IoU$\uparrow$ & CD$\times 10^{-2}$$\downarrow$ & F1$\uparrow$ & NC$\uparrow$     & IoU$\uparrow$ & CD$\downarrow$ & F1$\uparrow$ & NC$\uparrow$    & IoU$\uparrow$ & CD$\downarrow$ & F1$\uparrow$ & NC$\uparrow$ \\ 
\hline
SPSR                   & 0.333 & 3.225 & 0.523 & 0.852              & 0.204 & 0.438 & 0.267  & 0.755            & 0.234 & 0.105 & 0.245 & 0.841                  \\ 
LIG                   & 0.589 & 0.751 & 0.767 & 0.872              & 0.566 & 0.041 & 0.673 &  0.886            & 0.546 & 0.034 & 0.576 & 0.868                   \\ 
SGNN                    & 0.494 & 0.876 & 0.673 & 0.857                & 0.738 & 0.025& 0.804  & 0.919            & 0.441 & 0.029 & 0.471 & 0.867                   \\ 
ConvOcc                 & 0.600 & 0.779 & 0.765 & 0.913              & 0.565 & 0.037 & 0.667  & 0.905             & 0.419 & 0.034 & 0.420  & 0.859                    \\ 
IFNet                   & 0.777 & 0.420 & 0.937 & 0.923             & 0.779 & 0.028 & 0.832 & 0.918              & 0.575 & 0.029 & 0.607 & 0.866              \\ 
\hline
Ours                    & \textbf{0.783} & \textbf{0.377} & \textbf{0.947} & \textbf{0.938}               & \textbf{0.863} & \textbf{0.021} & \textbf{0.875} & \textbf{0.955}             & \textbf{0.710}  & \textbf{0.021} & \textbf{0.702}  & \textbf{0.917}                    \\
\hline
\end{tabular}
\caption{Reconstruction performance on the point cloud to surface reconstruction on ShapeNet, 3DFront, and Matterport3D.}
\label{tab:pcrecon}
\end{center}
\vspace{-0.50cm}
\end{table*}
}

\subsection{Ablations}
\begin{figure*}
	\centering
	\includegraphics[width=\linewidth]{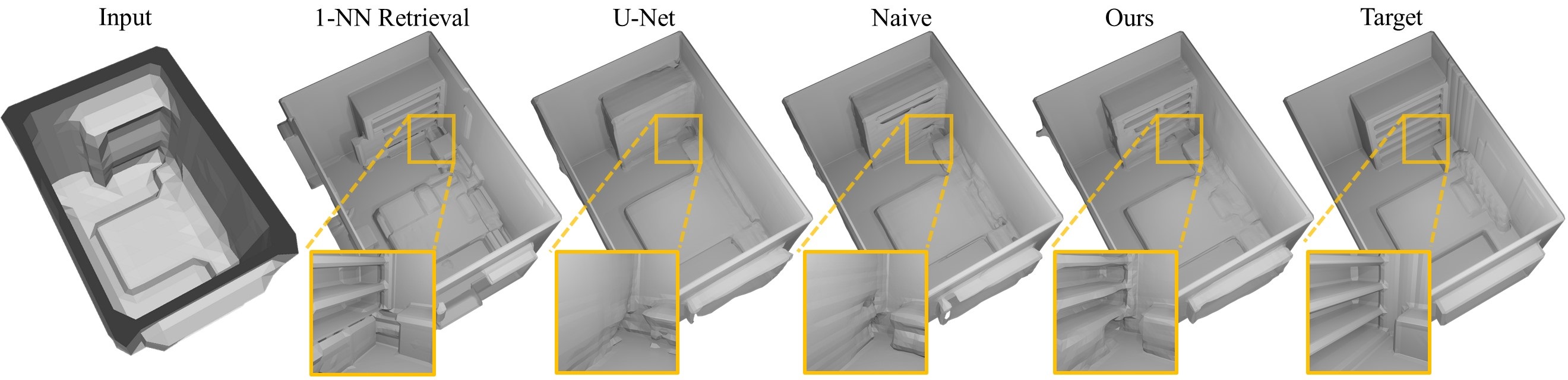}
	\caption{Qualitative evaluation of our method (\textit{Ours}) in comparison to $1^\mathrm{st}$ nearest neighbor retrieval (\textit{1-NN Retrieval}), a U-Net like network which doesn't use retrievals (\textit{U-Net}), and naive fusion of retrieved approximations during refinement (\textit{Naive}).}
	\label{fig:experiment_components}
	\vspace{-0.25cm}
\end{figure*}

\paragraph{Effect of retrieval and attention-based refinement.}
We evaluate the effect of our retrieval-based priors and attention-based refinement in Tab.~\ref{tab:ablation}.
We consider \emph{Retrieval} as the initial $1^\mathrm{st}$ nearest neighbor estimate provided by the retrieved scene data, \emph{U-Net} as a U-Net backbone styled similar to our refinement (and similar number of parameters to our refinement) but without using retrievals or attention as there are no retrievals to attend to, and \emph{Naive} to be our retrieval and refinement using concatenation of features instead of attention.
A visualization is shown in Fig.~\ref{fig:experiment_components}, with \emph{Retrieval} appearing disjoint between different retrieved chunks, \emph{U-Net} producing over-smoothed results, \emph{Naive} providing more details but still suffering from over-smoothing, and our method (with retrieval priors combined with attention-based refinement) producing the most consistent structure with local details defined.

{
\setlength{\tabcolsep}{4pt}
\begin{table}
\small
\resizebox{\linewidth}{!}{
\begin{tabular}{|l|l|l|l|l|l|l|l|l|} 
\hline
\multirow{2}{*}{Network}             & \multicolumn{4}{c|}{3D Super Resolution} & \multicolumn{4}{c|}{Surface Reconstruction}  \\ 
\cline{2-9}
                                     & IoU$\uparrow$ & CD$\downarrow$ & F1$\uparrow$ & NC$\uparrow$                    & IoU$\uparrow$ & CD$\downarrow$ & F1$\uparrow$ & NC$\uparrow$                            \\ 
\hline
Retrieval                            & 0.67 & 0.032 & 0.71 & 0.87                       & 0.70 & 0.028 & 0.75 & 0.88                              \\ 
U-Net                             & 0.68 & 0.029 & 0.77 & 0.91                       & 0.83 & 0.024 & 0.85 & 0.94                                \\ 
Naive                                & 0.71 & 0.028 & 0.77 & 0.91                      & 0.84 & 0.023 & 0.86 & \textbf{0.96}                               \\ 
\hline
Ours & \textbf{0.75} & \textbf{0.026} & \textbf{0.80} & \textbf{0.92}                       &  \textbf{0.86} & \textbf{0.021} & \textbf{0.88} & \textbf{0.96}                               \\
\hline
\end{tabular}
}
\vspace{-0.15cm}
\caption{Our attention based refinement performs better in comparison to not using any retrievals (\textit{U-Net}) or naivly fusing of retrievals during refinement (\textit{Naive}) on 3DFront dataset.}
\label{tab:ablation}
\vspace{-0.35cm}
\end{table}
}

\smallskip
\noindent \textbf{Effect of number of nearest neighbor retrievals.}
Tab.~\ref{tab:nn_ablation} shows the effect of increasing number of nearest neighbor retrievals used as a prior for the scene reconstruction.
With more nearest neighbors, the attention-based refinement has more candidates to select geometry from, improving  performance but with decreasing marginal gain.
\begin{table}
\centering
\small
\begin{tabular}{|l|l|l|l|l|c|} 
\hline
k & IoU$\uparrow$ & CD$\downarrow$ & F1$\uparrow$ & NC$\uparrow$ & \makecell{GPU\\memory}  \\ 
\hline
0 & 0.684 & 0.029 & 0.773 & 0.909 & 2.3G  \\ 
1 & 0.733 & 0.028 & 0.794 & 0.920  &  6.4G  \\ 
2 & 0.741 & 0.027 & 0.797 & \textbf{0.923}  &  7.9G \\ 
3 & 0.745 & 0.027 & 0.797 & 0.922  & 9.0G\\ 
\hline
4 & \textbf{0.751} & \textbf{0.027} & \textbf{0.801} & 0.922 & 9.9G \\
\hline
\end{tabular}
\vspace{-0.15cm}
\caption{Additional retrieval approximations help in improving refinement quality, at the cost of higher GPU memory utilization during training. Evaluation performed on 3D super-resolution task on 3DFront dataset.}
\label{tab:nn_ablation}
\vspace{-0.25cm}
\end{table}

\smallskip
\noindent \textbf{Extending the database during test time.}
Our approach can take advantage of new entries in the database without retraining.
Specifically, we conducted an experiment where we train on a subset of $8$ ShapeNet classes and evaluate it on other $5$ classes.
In Tab.~\ref{tab:unseen_classes}, we show that if we augment the database with chunks from the train set of the $5$ classes, our method improves without retraining by leveraging better retrievals.
Note that the other baselines would need to be retrained or refined to take advantage from new data.

\begin{table}
\centering
\small
\begin{tabular}{|l|l|l|l|l|l|} 
\hline
Variant & IoU$\uparrow$ & CD$\downarrow$ & F1$\uparrow$ & NC$\uparrow$ \\
\hline
Ours  & 0.478 & 0.034 & 0.601 & 0.811 \\
Ours (extended DB) & \textbf{0.579} & \textbf{0.029} & \textbf{0.743} & \textbf{0.825} \\
\hline
\end{tabular}
\vspace{-0.15cm}
\caption{Extending the database on ShapeNet subset during test time with additional new chunks leads to better reconstruction quality without the need of retraining.}
\label{tab:unseen_classes}
\vspace{-0.25cm}
\end{table}

\subsection{Implicit Reconstruction with a Database}
\begin{figure}
	\centering
	\includegraphics[width=\linewidth]{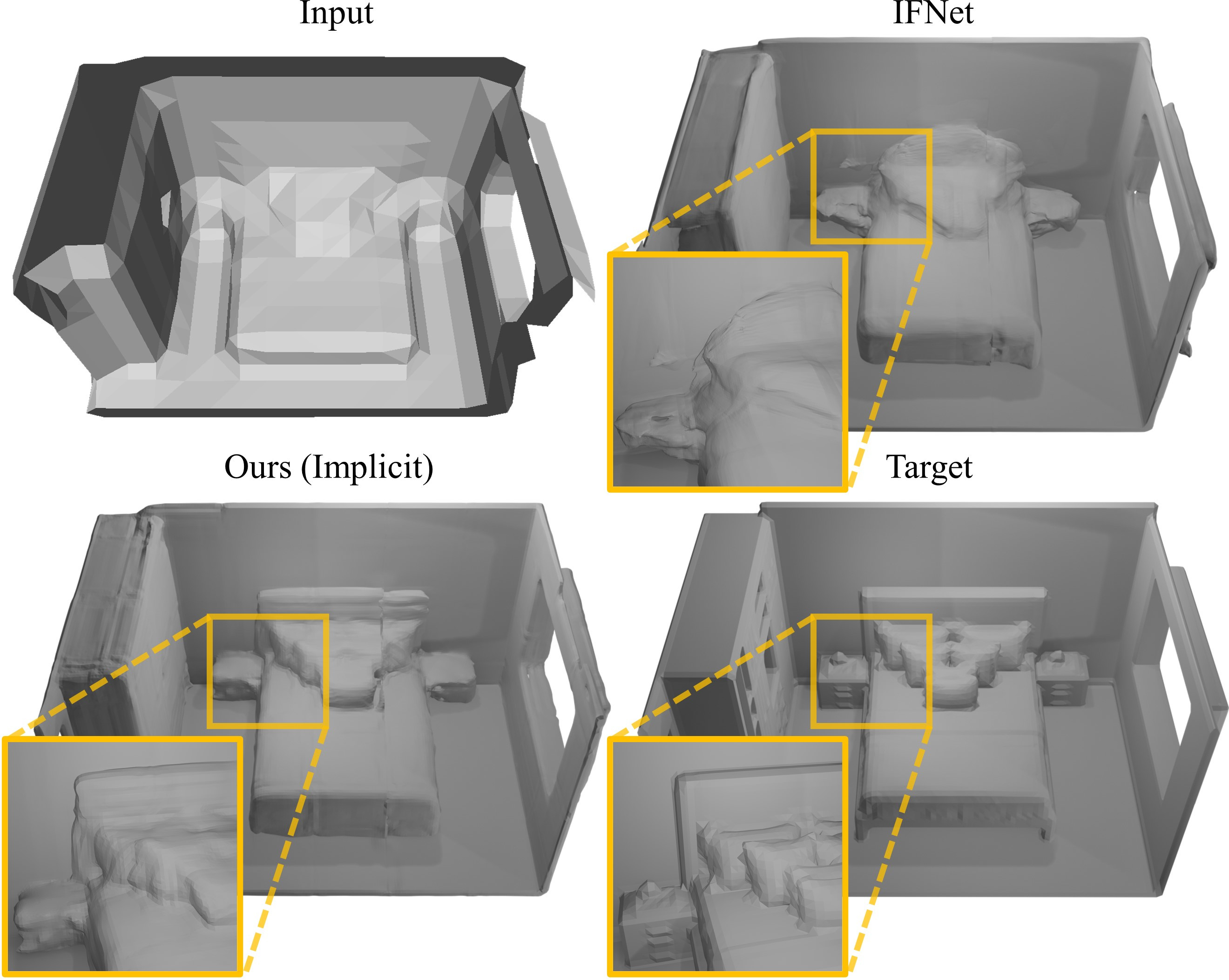}
	\vspace{-0.6cm}
	\caption{Qualitative evaluation of the implicit variant of our method on 3D super-resolution task on 3DFront dataset.}
	\label{fig:experiment_implicit}
	\vspace{-0.5cm}
\end{figure}

We can also apply our approach to implicit networks for 3D reconstruction by leveraging our retrieval estimates as an initial reconstruction for implicit-based refinement.
We thus incorporate our retrieval-based reconstruction with IFNet~\cite{chibane2020implicit}, maintaining our distance field database and incorporating the IFNet encoder (which also takes volumetric input) as well as IFNet decoder.
For additional architecture details, we refer to the appendix.
Tab.~\ref{tab:implicit_variant} shows that our retrieval-based reconstruction on 3DFront super-resolution task also helps to improve upon a learned implicit 3D reconstruction. Qualitative results are shown in Fig.~\ref{fig:experiment_implicit}.
\begin{table}
\centering
\small
\begin{tabular}{|l|l|l|l|l|} 
\hline
Method & IoU$\uparrow$ & CD$\downarrow$ & F1$\uparrow$ & NC$\uparrow$ \\ 
\hline
IFNet & 0.639 & 0.041 & 0.736 & 0.878 \\ 
Ours (Implicit) & {\bf 0.687} & {\bf 0.038} & {\bf 0.766} & {\bf 0.897} \\ 
\hline
\end{tabular}
\vspace{-0.15cm}
\caption{Performance of an implicit variant of method that extends IFNet. Evaluated on 3D super-resolution task on 3DFront dataset.}
\label{tab:implicit_variant}
\vspace{-0.5cm}
\end{table}

\paragraph{Limitations.}
Our approach to leverage high-quality train scene data for 3D reconstruction shows notable improvements from state of the art; however, several limitations remain.
Our retrieval estimates and refinement operate in two separate stages without gradient propagation from the final reconstruction to the retrieved scene data, resulting in possible suboptimal retrieval where the refinement must compensate more; developing a differentiable $k$-NN retrieval~\cite{plotz2018neural,tseng2020retrievegan} for refinement could bridge these disconnects.
Additionally, by constructing our retrieval dictionary from chunks of train scenes, our dictionary size can grow quite large; we believe a learned dictionary could help to construct the most informative characteristics to be leveraged for reconstruction.
We refer to the appendix for visualisation and discussion of limitations and failure cases.

\section{Conclusion}
We introduce a new approach to geometric scene reconstruction, leveraging learned retrieval of train scene data as a basis for attention-based refinement to produce more globally consistent reconstructions while maintaining local detail.
By first constructing a reconstruction estimate composed of train scene chunks, we can learn to propagate desired geometric properties inherent in existing scene data such as clean structures and local detail, through our patch-based attention in the reconstruction refinement.
This produces more accurate scene reconstructions from low-resolution or point cloud input, and opens up exciting avenues for exploiting constructed or even learned dictionaries for challenging generative 3D tasks.

\paragraph{Acknowledgements.}
{
\small
This work was supported by the Bavarian State Ministry of Science and the Arts  coordinated by the Bavarian Research Institute for Digital Transformation (bidt), a TUM-IAS Rudolf M{\"o}{\ss}bauer Fellowship, an NVidia Professorship Award, the ERC Starting Grant Scan2CAD (804724), and the German Research Foundation (DFG) Grant Making Machine Learning on Static and Dynamic 3D Data Practical. Apple was not involved in the evaluations and implementation of the code. 
}

\newpage
{\small
\bibliographystyle{ieee_fullname}
\bibliography{bib}
}

\clearpage
\newpage
\begin{appendix}
In this appendix, we discuss additional experiments that we conducted with our neural 3D scene reconstruction method \textit{RetrievalFuse} (Sec.~\ref{sec:appendix_evaluation}).
Specifically, we show additional ablation studies and results for both the 3D super-resolution and surface reconstruction.
We also provide implementation details of our method and the used baselines (Section~\ref{sec:appendix_impl}), as well as our data generation (Sec.~\ref{sec:appendix_datagen}).
We conclude with a discussion about limitations.

\section{Implementation Details}
\label{sec:appendix_impl}

\begin{figure}[b!]
	\centering
	\includegraphics[width=\linewidth]{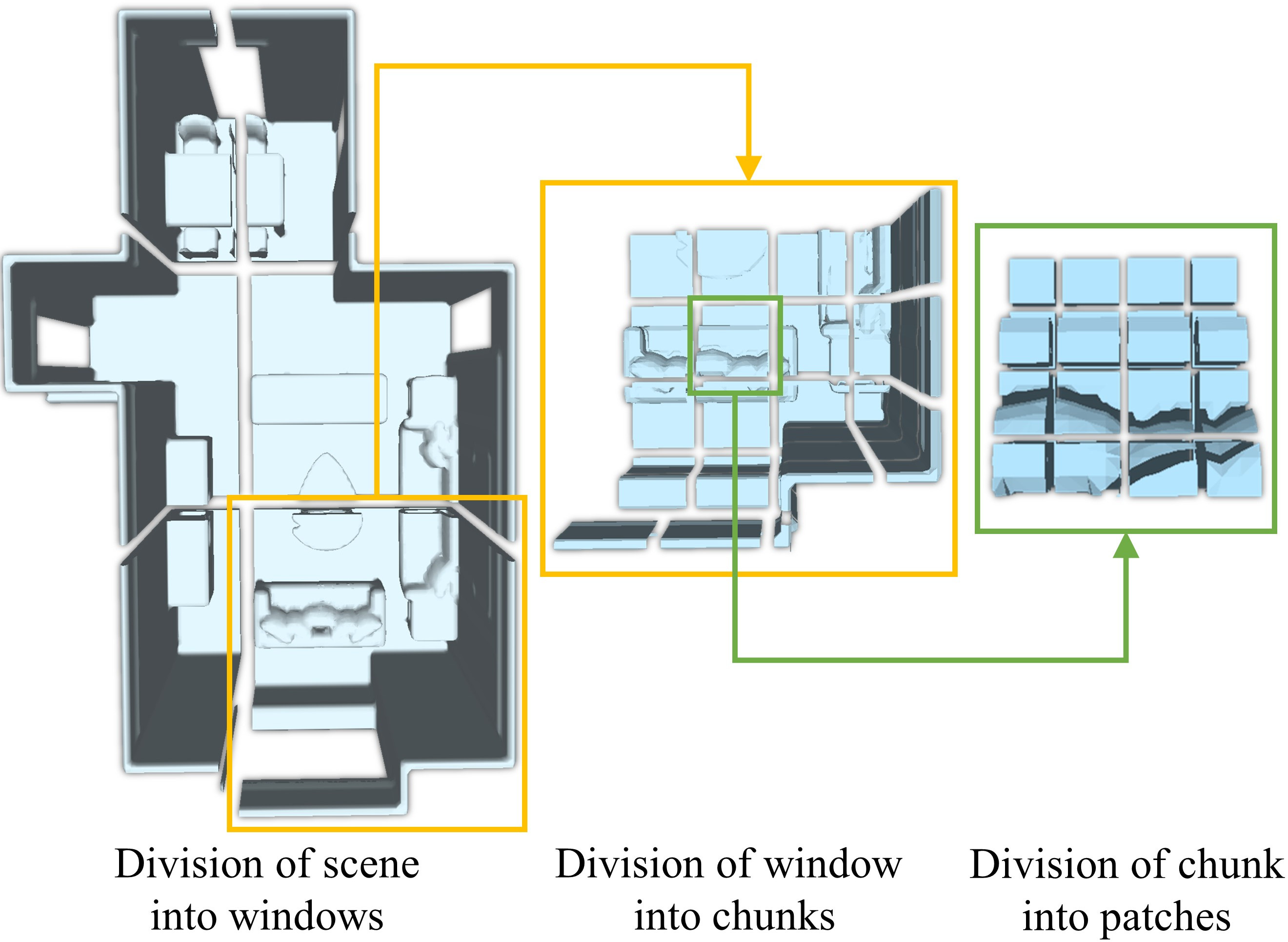}
	\caption{In our experiments, we use $64^3$ target chunks for target geometry, and larger scenes work in a sliding window fashion (left). The retrieval candidates are $16^3$ chunks (middle), and attention-based blending works on $4^3$ patches (right).}
	\label{fig:level_of_operation}
\end{figure}

\paragraph{Levels of Operation for Scene Reconstruction.}
Fig.~\ref{fig:level_of_operation} shows the different levels of operation at which our method operates on to reconstruct a 3D scene. 
Larger scenes are split into fixed size windows, chunk retrievals are made on smaller sized chunks for more expressability, and attention-based blending works on yet smaller sized patches to allow the method to choose among different retrievals at a finer detail. 

\begin{figure*}
	\centering
	\includegraphics[width=\linewidth]{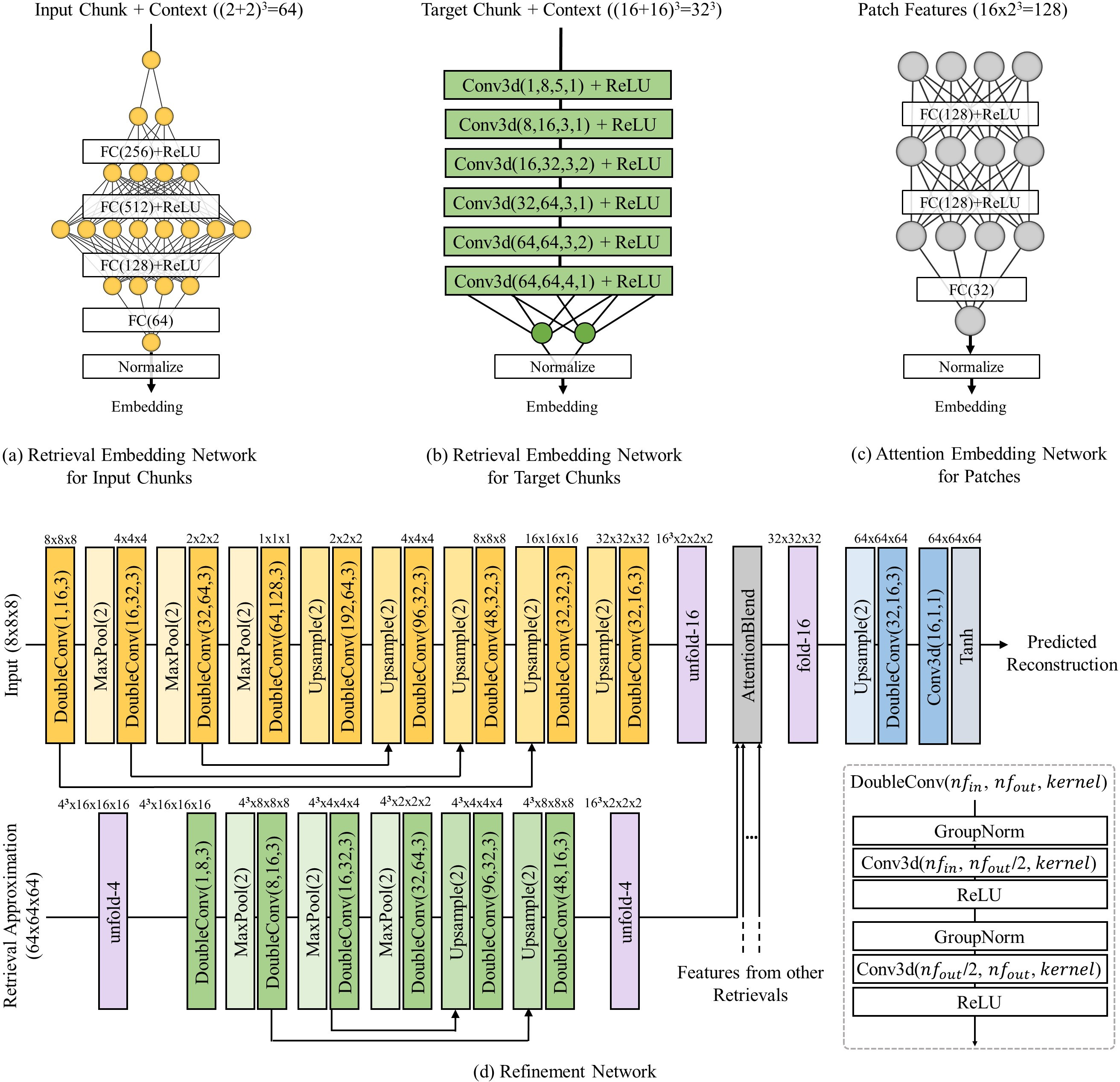}
	\caption{Network architecture used in our 3D super-resolution experiments. Convolution parameters are given as (input features, output feature, kernel size, stride), with default stride of 1 if not specified. Array of circles represent fully connected (FC) layers. 
	For the task of point cloud to surface reconstruction, the input chunk embedding network is a convolutional layer instead of MLP with a fully connected layer at the end on account of larger input chunk size (since input is a $128^3$ grid for surface reconstruction in comparison to $8^3$ grid for super-resolution, we use a chunk size of $32^3$ for inputs there). Additionally, the input feature extractor is deeper for point cloud to surface reconstruction on account on bigger input grid.}
	\label{fig:architecture_refine}
\end{figure*}

\paragraph{Network Architecture.}
Fig.~\ref{fig:architecture_refine} details the architecture of our networks for 3D super-resolution task. All networks are implemented in PyTorch~\cite{NEURIPS2019_9015}.

\paragraph{Inference Time and Number of Parameters.}
We report the number of trainable parameters and the inference time for our method (both retrieval and refinement stage) along with that of the baselines in Tab.~\ref{tab:inference_params} for the 3D super-resolution task. All runtimes are reported on a machine with Intel(R) Xeon(R) Gold 6240 CPU @ 2.60GHz processor with an NVIDIA 2080Ti GPU. We use FLANN~\cite{muja2009fast} to speed up nearest neighbor lookups from the database. Our retrieval inference time is significantly higher than refinement due to multiple disk reads to retrieve chunks ($=$ number of chunks $\times$ number of retrievals). To avoid this overhead during training, once the retrieval networks have been trained, we preprocess the entire training set to extract retrievals before starting refinement stage training.
{
\begin{table}
    \centering
    \small
    \begin{tabular}{|l|l|l|} 
    \hline
    Method & Inference Time (s) & \# Parameters ($\times10^6$) \\
    \hline
    SGNN~\cite{dai2020sg} & 2.297 & 0.64\\
    ConvOcc~\cite{peng2020convolutional} & 1.707 & 1.04\\
    IFNet~\cite{chibane2020implicit} & 0.708 & 2.95\\
    Ours (Retrieval) & 0.784 & 0.77\\
    Ours (Refinement) & 0.012 & 1.49\\
    \hline
    \end{tabular}
    \caption{Comparison of inference time and number of trainable parameters on the 3D super-resolution task.}
    \vspace{-0.25cm}
    \label{tab:inference_params}
\end{table}
}

{
\setlength{\tabcolsep}{5pt}
\begin{table}
    \centering
    \small
    \resizebox{\linewidth}{!}{
    \begin{tabular}{|l|l|l|l|l|l|l|l|} 
    \hline
    \multirow{2}{*}{ \makecell{Chunk\\side (m)}} & \multicolumn{4}{c|}{Retrieval} & \multicolumn{3}{c|}{Refinement}  \\ 
    \cline{2-8}
                                & IoU$\uparrow$ & CD$\downarrow$ & NC$\uparrow$ & Entries & IoU$\uparrow$ & CD$\downarrow$ & NC$\uparrow$    \\ 
    \hline
    3.467                            & 0.53 & 0.074 &  0.72 & 43092  & 0.71 & 0.029 & 0.91   \\
    1.733                            & 0.60 & 0.041 & 0.85  &  344249 & 0.72 & 0.028 & 0.91 \\
    0.867                            & \textbf{0.67} & \textbf{0.033} & \textbf{0.87} & 2093592 & \textbf{0.75} & \textbf{0.026} & \textbf{0.92}  \\
    \hline
    \end{tabular}
    }
    \caption{Smaller sized chunk retrievals improve the performance of both retrieval and refinement, although at cost of a larger database. Evaluation performed on 3D super-resolution task on 3DFront dataset.}
    \label{tab:patchsize_ablation}
\end{table}
}

\begin{table}
    \centering
    \small
    \begin{tabular}{|l|l|l|l|l|l|} 
        \hline
        Variant & IoU$\uparrow$ & CD$\downarrow$ & F1$\uparrow$ & NC$\uparrow$ \\
        \thickhline
        Retrieval & 0.364 & 0.781 & 0.525 & 0.708 \\
        Backbone & 0.463 & 0.647 & 0.602 & 0.813 \\
        Naive & 0.432 & 0.684 & 0.576 & 0.798 \\
        Ours  & 0.478 & 0.635 & 0.601 & 0.811 \\
        \hline
    \end{tabular}
    \caption{In case of suboptimal retrievals, our method does not provide significant improvement over the backbone reconstruction quality. However, it is more robust to bad retrievals compared to a naive blending of retrieval features with input features. Networks trained on a ShapeNet subset with 8 classes and evaluated on a disjoined subset with 5 classes.}
    \label{tab:appendix_unseen_classes}
\end{table}

\begin{table}
    \centering
    \small
    \begin{tabular}{|l|l|l|l|} 
        \hline
        \# Train Scenes & IoU$\uparrow$ & CD$\downarrow$ & F1$\uparrow$ \\ 
        \hline
        3750 ~~~(25\%) & 0.711 & 0.0283 & 0.784 \\ 
        7500 ~~~(50\%) & 0.728 & 0.0275 & 0.791 \\ 
        11250 ~(75\%) & 0.741 & 0.0269 & 0.796 \\ 
        15000 ~(100\%) & 0.751 &  0.0265 & 0.801 \\
        \hline
    \end{tabular}
    \vspace{0.15cm}
    \caption{Ablation study w.r.t. the number of train scenes, evaluated on the 3D super-resolution task using the 3DFront dataset.}
    \label{tab:num_scene}
\end{table}

\subsection{IFNet-based RetrievalFuse}

\begin{figure}
	\centering
	\includegraphics[width=\linewidth]{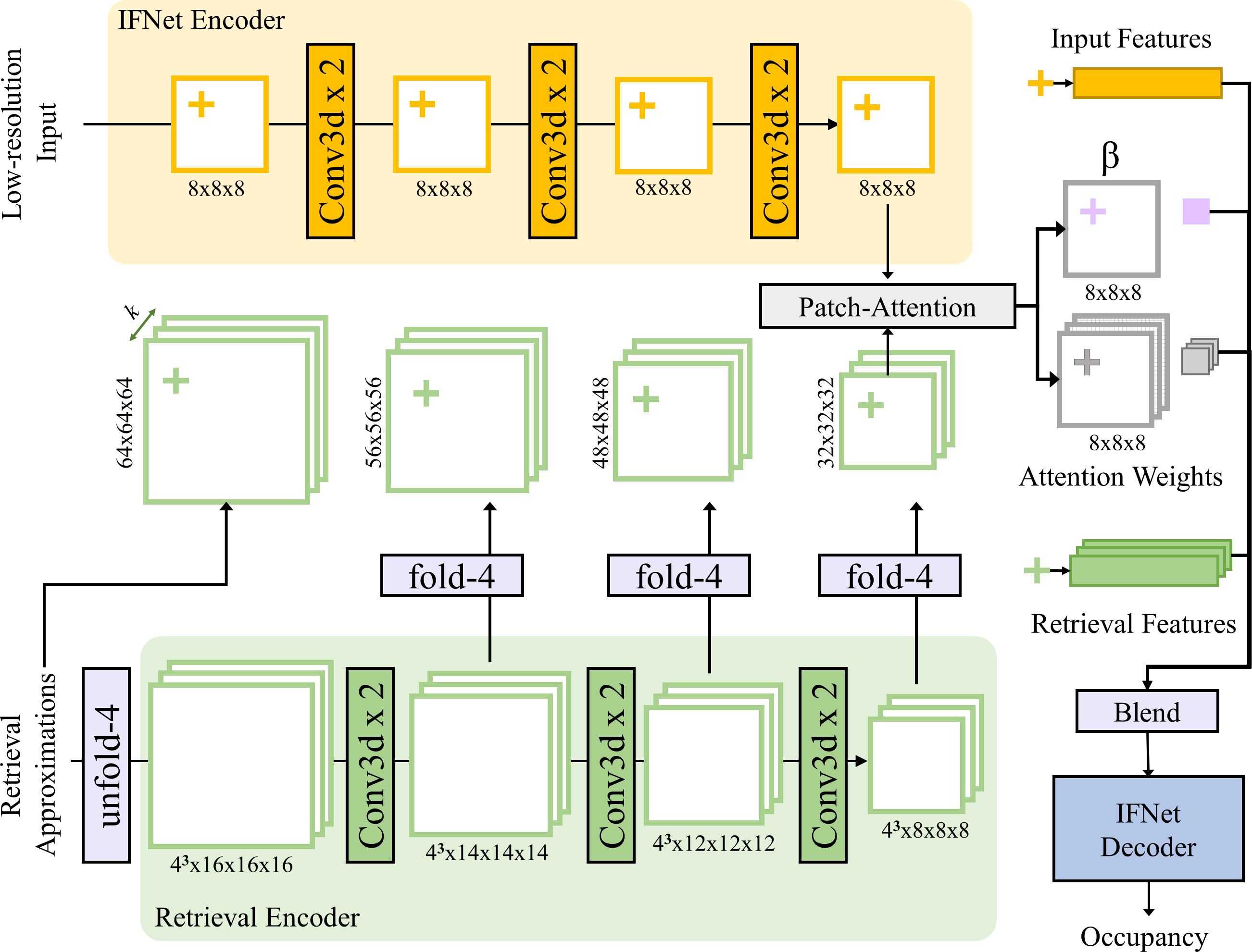}
	\caption{Integration of our RetrievalFuse approach to the implicit network of IFNet~\cite{chibane2020implicit}. 
	We use IFNet's encoder as the input feature encoder and their decoder as the implicit decoder. Additionally, we use a retrieval encoder similar to the IFNet encoder for obtaining features for the retrieval approximations. Further, a patch attention layer computes a blend coefficient grid and attention weight grid. For a given query point in space, features are sampled from input feature grids, retrieval feature grids. A blend coefficient value and attention weights are sampled from the blend coefficient grid and attention weight grid at the queried point. The sampled input features and retrieval features are blended based on these valued and finally decoded to an occupancy value by the IFNet decoder.}
	\label{fig:architecture implicit}
\end{figure}

To demonstrate the wide applicability of our method, we also demonstrate our approach integrated into the implicit-based reconstruction of IFNet~\cite{chibane2020implicit} to leverage our retrieved approximations.
We keep the IFNet encoder and decoder unmodified, and add an additional retrieval encoder for processing the retrieved reconstruction approximations.
This retrieval encoder is based on the original IFNet encoder, and works with chunks from the retrievals.
For a given point in space, features sampled at the point from feature volumes at different levels of the input encoder make up the input features.
Features sampled from the retrieval features volumes at this point for each of the $k$ retrievals make up the retrieval features.
Next, based on the feature volume at last layer of input and retrieval encoder, a blending coefficient grid and an attention weight grid is obtained.
To obtain these, the $8\times8\times8$ input feature volume and the $32\times32\times32$ retrieval feature volume are interpreted as 512 patch volumes of shape $1\times1\times1$ and $4\times4\times4$ respectively.
These input and corresponding retrieval patch volumes are mapped to a shared embedding space, from which we can get the blending coefficient (Eq.~6, main paper) and attention weights (Eq.~4, main paper).
Once we have the blending coefficient grid and attention weight grid, we can sample their values at the queried point.
Finally we blend the sampled input features and the sampled $k$ retrieved features (Eq.~5, main paper) to give the blended feature that is decoded by the IFNet decoder.
%

\subsection{Baselines}
We use the official implementations provided by the authors of IFNet~\cite{chibane2020implicit}, Convolutional Occupancy Networks~\cite{mescheder2019occupancy}, SGNN~\cite{dai2020sg}, Local Implicit Grids~\cite{jiang2020local} and Screened Poisson Reconstruction~\cite{kazhdan2013screened} in our experiments.
For 3D super-resolution experiments, the methods are provided with low-resolution distance field grids as inputs instead of voxel grid inputs.
In particular, for IFNet we use the \textit{ShapeNet32Vox} model for 3D super-resolution.
For surface reconstruction from point clouds for IFNet, the $128^3$ discretized point cloud is used with the \textit{ShapeNetPoints} model.
For Convolutional Occupancy Networks we use the $32^3$ \textit{voxel simple encoder} for 3D super-resolution, and a $64^3$ \textit{point net local pool} encoder for point cloud surface reconstruction.
For SGNN, we use a $64^3$ resolution with nearest-neighbor upsampling to a $64^3$ grid for the input.
For Local Implicit Grids we found that the part sizes $0.25\times$ shape size for ShapeNet and $0.35\times$ window size for 3DFront and Matterport3D worked best at the sparsity of the input point cloud.


\section{Data Generation and Evaluation Metrics}
\label{sec:appendix_datagen}

\paragraph{Data generation.}
%
As specified in the main paper, the targets for both 3D super-resolution and surface reconstruction from point cloud tasks are $64^3$ distance field grids.
Training and inference on larger scenes is done in a sliding window manner with a window stride of 64.
We use SDFGen\footnote{\href{https://github.com/christopherbatty/SDFGen}{https://github.com/christopherbatty/SDFGen}} to generate these distance field targets.
Low-resolution distance field inputs are generated in a similar manner at a coarser resolution.
Point cloud samples for surface reconstruction task are generated as random samples on the surface of meshes generated from target distance fields. 

For IFNet~\cite{chibane2020implicit}, Convolutional Occupancy Networks~\cite{mescheder2019occupancy}, and our implicit variant, all of which need supervision in the form of points along with their occupancies, we first extract meshes from the target distance fields using the marching cubes algorithm~\cite{lorensen1987marching}.
These meshes are then made watertight using \textit{implicit waterproofing}~\cite{chibane2020implicit} from which points and their occupancies are finally sampled.
SGNN is provided the same inputs and targets as ours for training, with the respective inputs upsampled to match the target $64^3$ resolution grid.
Local Implicit Grids~\cite{jiang2020local} is trained on ShapeNet, and Screened Poisson Reconstruction~\cite{kazhdan2013screened} does not require training; however, both methods are provided high-resolution normals to obtain oriented point clouds as inputs.

\paragraph{Evaluation Metrics.}
%
We follow the definition and implementations of Chamfer $\ell_1$ Distance, Normal Consistency, and F-Score from \cite{peng2020convolutional}.
Specifically, Chamfer $\ell_1$ Distance (CD) is defined as:
\begin{equation*}
    \small
    \begin{split}
        \mathrm{CD}(\mathcal{M}_{pred}, \mathcal{M}_{gt}) = \frac{1}{2}(\mathrm{Acc}(\mathcal{M}_{pred}, \mathcal{M}_{gt}) \\ +  \mathrm{Comp}(\mathcal{M}_{pred}, \mathcal{M}_{gt}))
    \end{split}
\end{equation*}
where $\mathcal{M}_{pred}$ and $\mathcal{M}_{gt}$ are the predicted and target meshes (obtained by running marching cubes on predicted and target distance fields).
$Acc(.)$ and $Comp(.)$ are accuracy and completeness given as:
\small
\begin{equation*}
        \mathrm{Acc}(\mathcal{M}_{pred}, \mathcal{M}_{gt}) = \frac{1}{\left|\partial\mathcal{M}_{pred}\right|}\int_{\partial\mathcal{M}_{pred}}\min_{\mathbf{q}\in \partial\mathcal{M}_{gt}}\norm{\mathbf{p} - \mathbf{q}}\mathrm{d}\mathbf{p},
\end{equation*}
\normalsize
and
\small
\begin{equation*}
    \mathrm{Comp}(\mathcal{M}_{pred}, \mathcal{M}_{gt}) = \frac{1}{\left|\partial\mathcal{M}_{gt}\right|}\int_{\partial\mathcal{M}_{gt}}\min_{\mathbf{p}\in \partial\mathcal{M}_{pred}}\norm{\mathbf{p} - \mathbf{q}}\mathrm{d}\mathbf{q}
\end{equation*}
\normalsize
with $\partial\mathcal{M}_{pred}$ and $\partial\mathcal{M}_{gt}$ denoting the surfaces of the meshes.
Normal Consistency (NC) is defined as:
\begin{align*}
    \footnotesize
    \begin{split}
    \mathrm{NC}(\mathcal{M}_{pred}, \mathcal{M}_{gt}) &= \frac{1}{2\left|\partial\mathcal{M}_{pred}\right|}\int_{\partial\mathcal{M}_{pred}}\abs{n(\mathbf{p}) \cdot n(\mathrm{proj}_2(\mathbf{p}))}\mathrm{d}\mathbf{p} \\ &+
    \frac{1}{2\left|\partial\mathcal{M}_{gt}\right|}\int_{\partial\mathcal{M}_{gt}}\abs{n(\mathbf{q}) \cdot n(\mathrm{proj}_1(\mathbf{q}))}\mathrm{d}\mathbf{q}
    \end{split}
\end{align*}
where $(.)$ indicates inner product, $n(\mathbf{p})$ and $n(\mathbf{q})$ are the unit normal vectors on the mesh surface, and $\mathrm{proj}_2(\mathbf{p})$ and $\mathrm{proj}_1(\mathbf{q})$ are projections of $\mathbf{p}$ and $\mathbf{q}$ onto mesh surfaces $\partial\mathcal{M}_{pred}$ and $\partial\mathcal{M}_{gt}$ respectively.
F-Score \cite{tatarchenko2019single} is defined as the harmonic mean of precision and recall, where recall is fraction of points on $\mathcal{M}_{gt}$ that lie within a certain distance to $\mathcal{M}_{pred}$, and precision is the fraction of points on $\mathcal{M}_{pred}$ that lie within a certain distance to $\mathcal{M}_{gt}$.
For calculating the volumetric IoU, we first voxelize the meshes $\mathcal{M}_{gt}$ and $\mathcal{M}_{pred}$ with voxel sizes of $0.054$m for 3DFront, $0.0375$m for Matterport3D, and resolutions $64^3$ for ShapeNet.
The IoU is then given as:
\begin{equation*}
    \small
    \mathrm{IoU} = \frac{\mathrm{Voxels}(\mathcal{M}_{pred}) \cap \mathrm{Voxels}(\mathcal{M}_{gt})}{\mathrm{Voxels}(\mathcal{M}_{pred}) \cup \mathrm{Voxels}(\mathcal{M}_{gt})}
\end{equation*}

\begin{figure*}
	\centering
	\includegraphics[width=\linewidth]{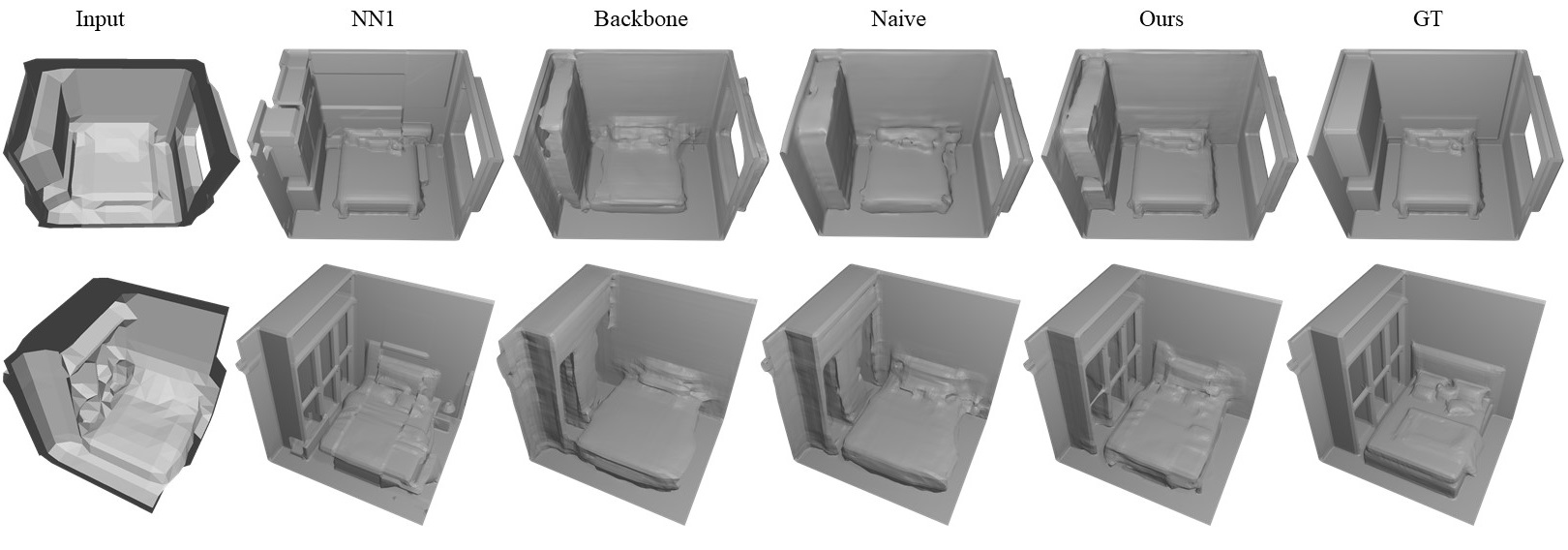}
	\caption{Additional qualitative evaluation of our method (\textit{Ours}) in comparison to $1^\mathrm{st}$ nearest neighbor retrieval (\textit{1-NN Retrieval}), our refinement network without retrievals (\textit{Backbone}) and naive fusion of retrieved approximations during refinement (\textit{Naive}).}
	\label{fig:appendix_components}
\end{figure*}


\section{Additional Evaluation}
\label{sec:appendix_evaluation}

\subsection{Ablation Studies}

\begin{figure*}
	\centering
	\includegraphics[width=\linewidth]{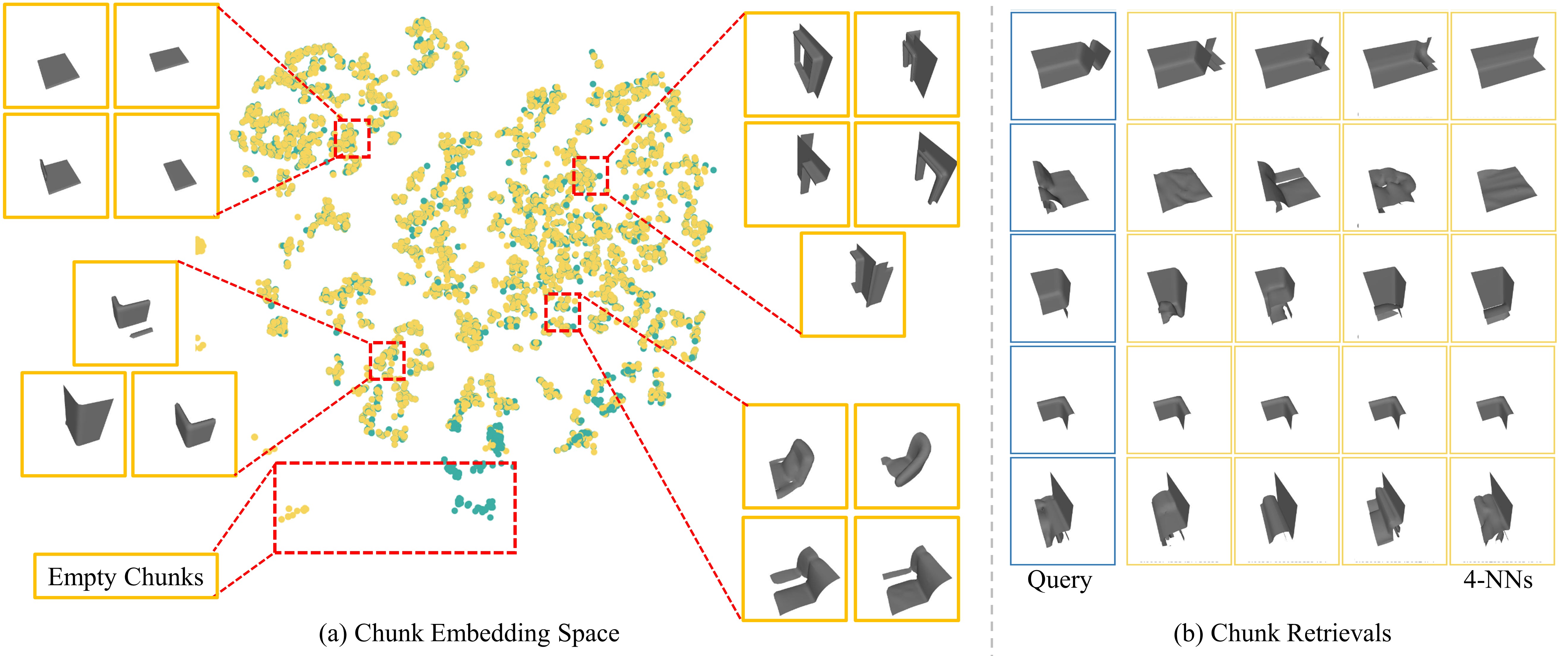}
	\caption{(a) Chunk embedding space visualized for 5000 chunks from 3DFront test set. This embedding space used for retrievals from the database by projecting an input chunk into this space (visualized as green dots) and retrieving k-nearest database chunks (visualized by yellow dots) from it. (b) Input queries and their corresponding 4 nearest neighbors from the embedding space. For the sake of visual clarity, input queries are visualized as their corresponding ground truth reconstruction.}
	\label{fig:latent_space}
\end{figure*}

\paragraph{Chunk Embedding Space Visualization.}
Fig.~\ref{fig:latent_space} visualizes the embedding space used for retrieving chunks from our database.
Chunks with similar geometry end up lying closer in this space.

\paragraph{Effect of retrieved chunk size on the performance of our method.}
Tab.~\ref{tab:patchsize_ablation} evaluates our method with retrieval approximations of different chunk sizes for retrieval.
A chunk size that is too large cannot effectively capture the diversity of various scene arrangements, while smaller sizes can represent a wider variety of geometry, at the cost of an increased database size.

\paragraph{Effect of number of training scenes used for creating the database}
Tab.~\ref{tab:num_scene} shows the effect of number of chunks in the database on our method's performance. Availability of a wider variety of chunks helps reconstruction.

\subsection{Additional Qualitative Results}
\begin{figure*}
	\centering
	\includegraphics[width=\linewidth]{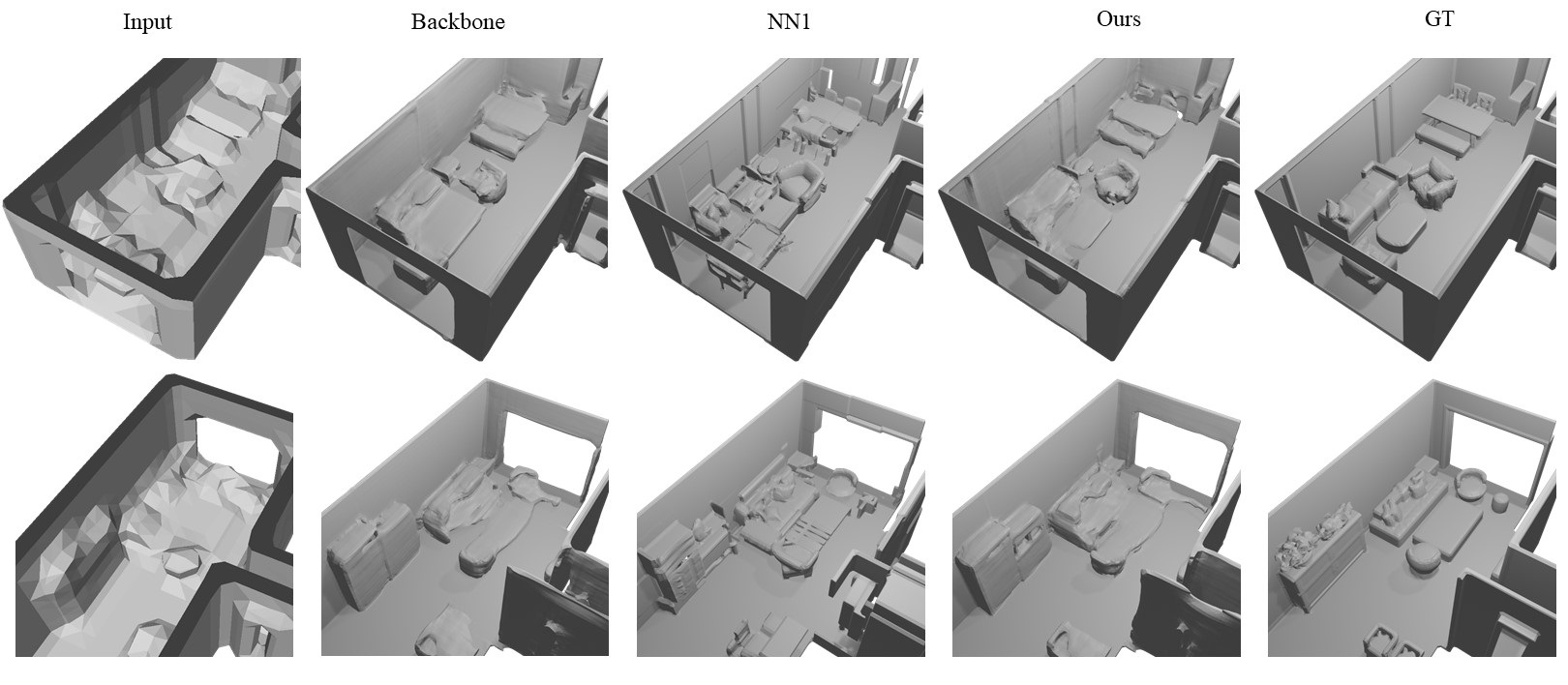}
	\caption{Suboptimal retrievals do not improve results significantly over our Backbone network. However, reconstruction produced are also not degraded due to subobtimal retrievals. Qualitative results from 3DFront super-resolution task.}
	\label{fig:bad_retrieval_3dfront}
\end{figure*}

\begin{figure*}
	\centering
	\includegraphics[width=\linewidth]{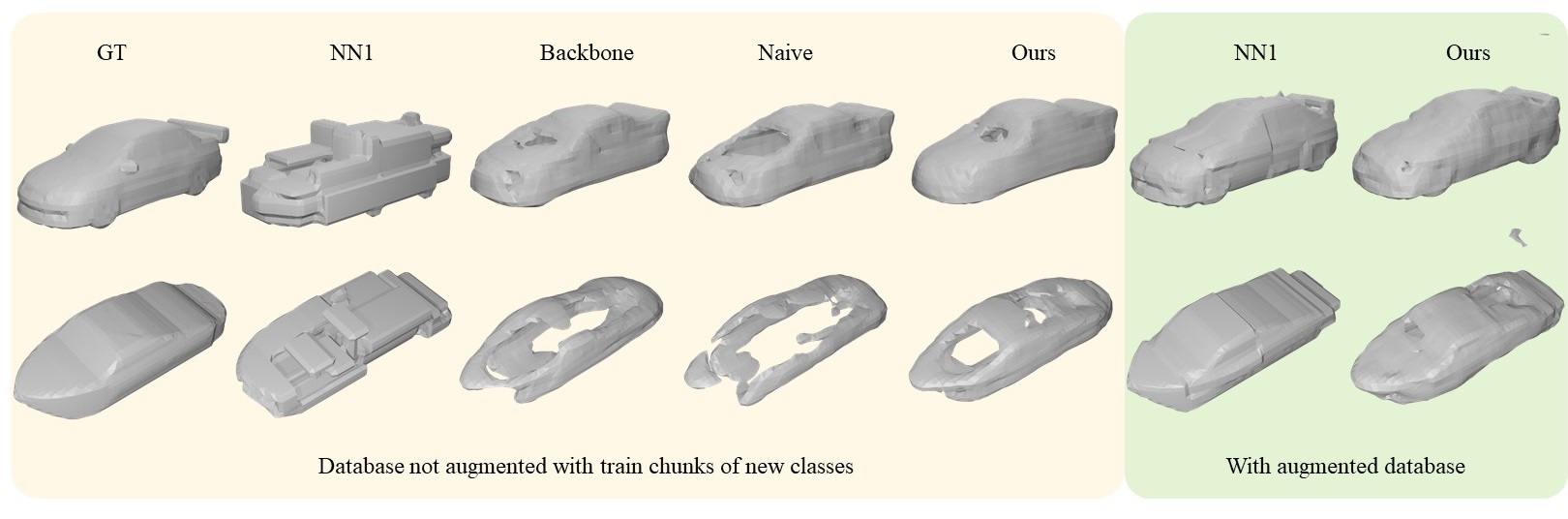}
	\caption{(Left) Suboptimal retrievals (NN1) when the our method is trained on a ShapeNet subset of 8 classes and evaluated on another 5 classes. The database contains chunks only from the original 8 classes. In this case, the suboptimal retrievals don't help the reconstruction, and the quality of reconstruction does not significantly improve over our backbone network. However, in contrast to naive fusion of retrieval features, our reconstruction quality does not degrade over the backbone. (Right) If the database if augmented with new chunks from train set of the new 5 classes, the reconstruction quality visibly improves without retraining.}
	\label{fig:shapenet_transfer_robust}
\end{figure*}

\begin{figure*}
	\centering
	\includegraphics[width=\linewidth]{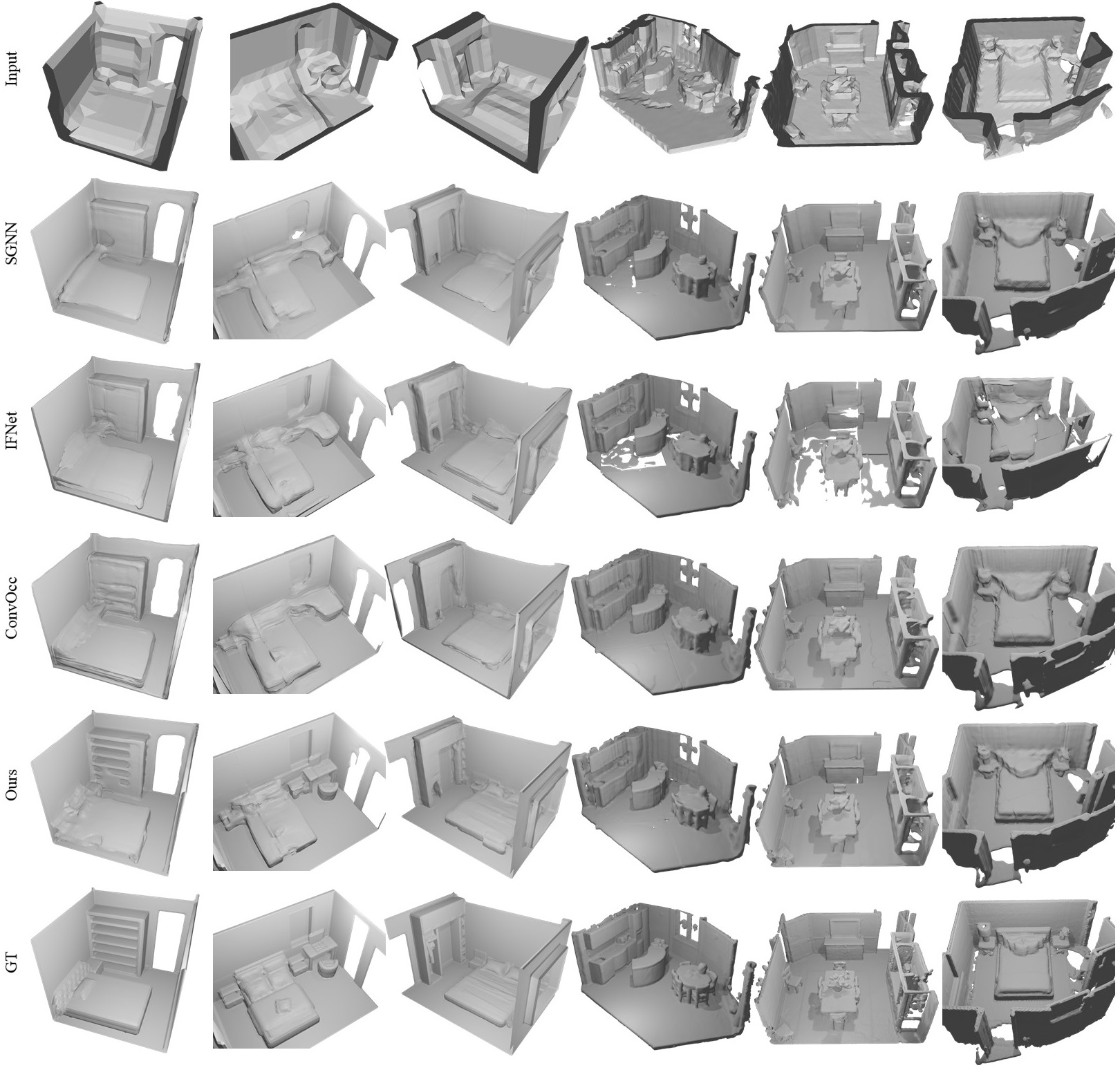}
	\caption{Additional qualitative results on 3DFront (left three) and Matterport3D (right three) on 3D super-resolution task.}
	\label{fig:appendix_superresolution}
\end{figure*}

\begin{figure*}
	\centering
	\includegraphics[width=\linewidth]{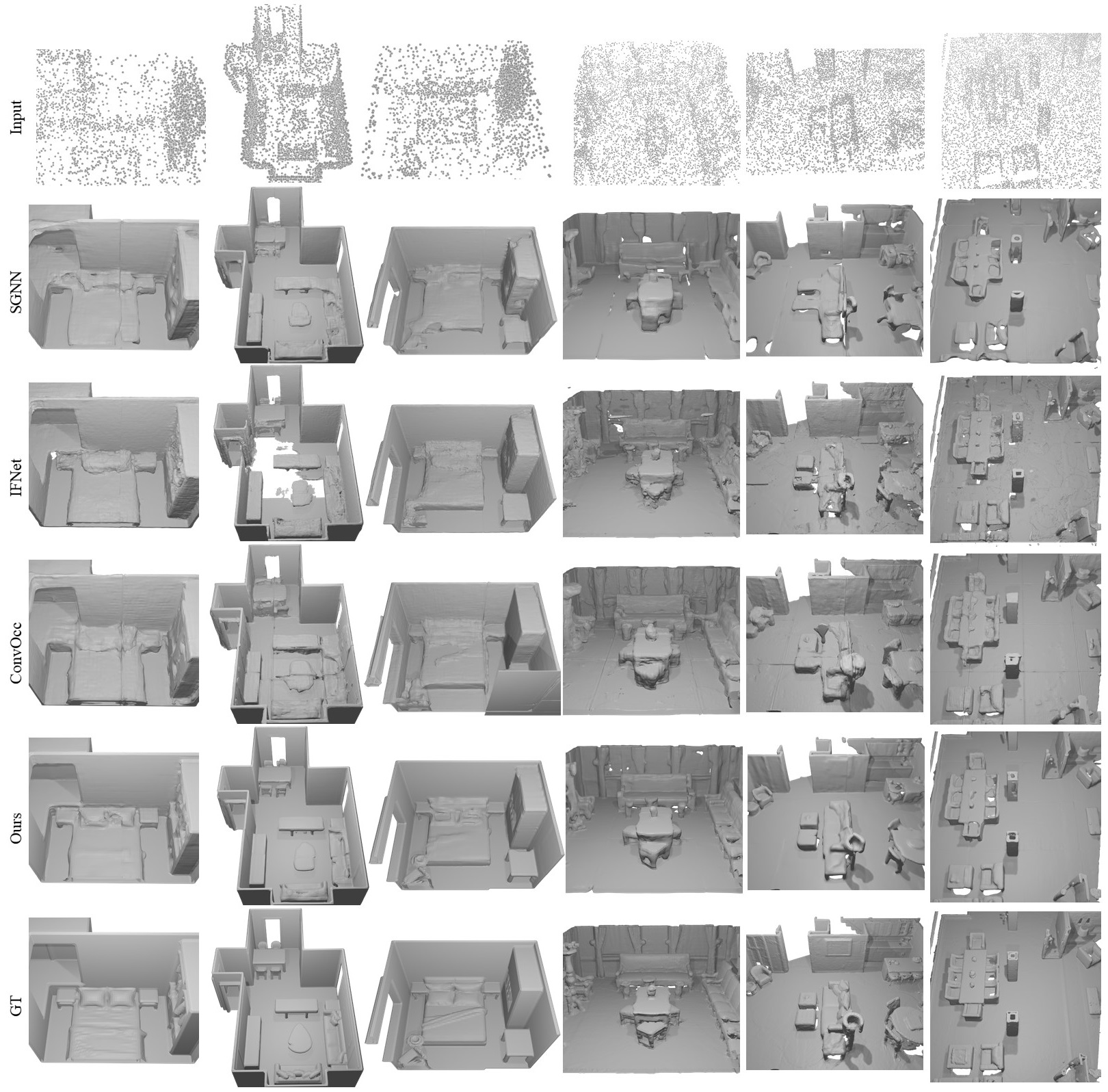}
	\caption{Additional qualitative results on 3DFront (left three) and Matterport3D (right three) on point cloud to surface reconstruction task.}
	\label{fig:appendix_surface_reconstruction}
\end{figure*}

\begin{figure*}
	\centering
	\includegraphics[width=\linewidth]{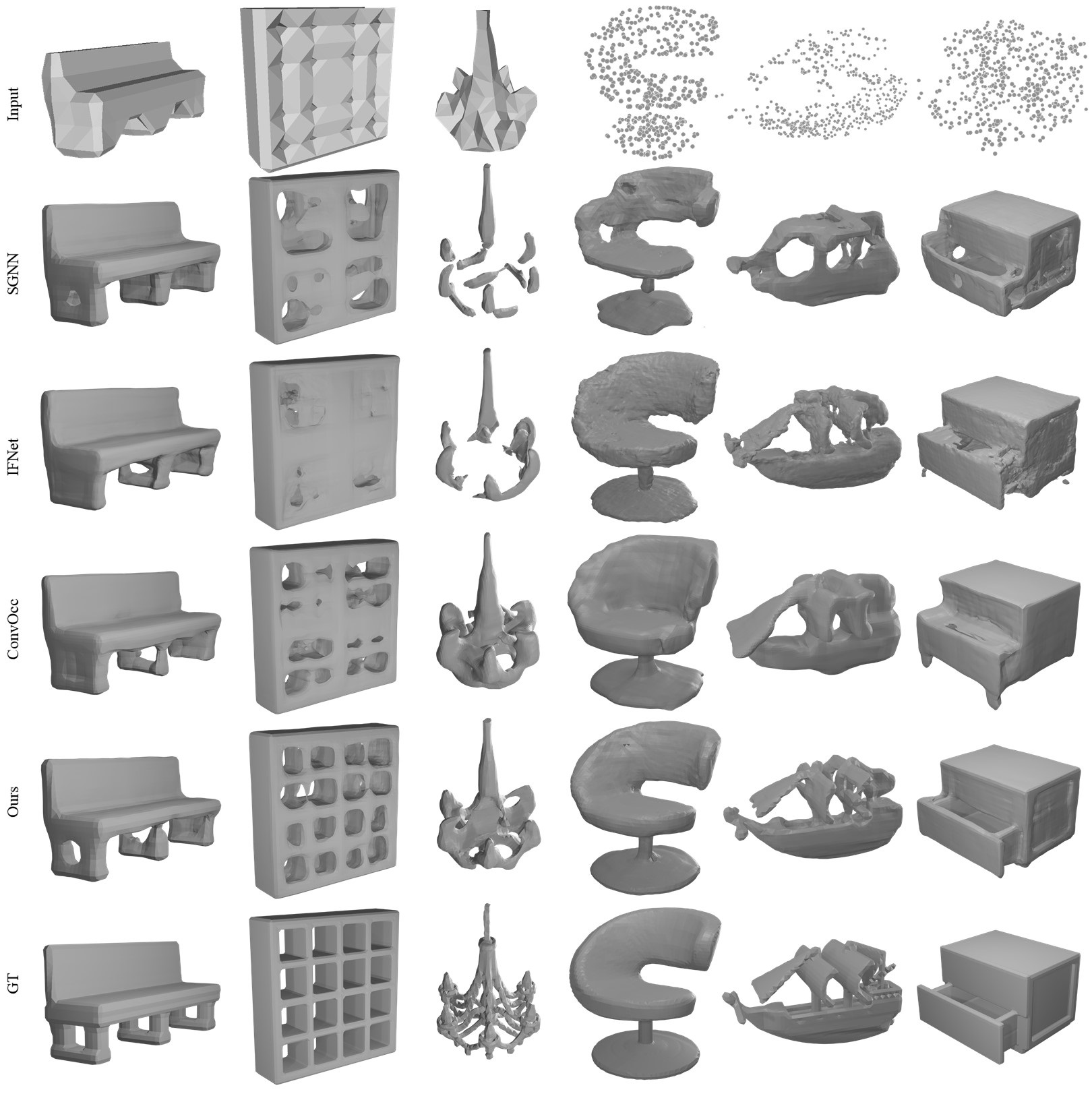}
	\caption{Qualitative results on ShapeNet dataset on 3D super-resolution (left three) and point cloud to surface reconstruction (right three) tasks.}
	\label{fig:appendix_shapenet}
\end{figure*}

We provide additional qualitative evaluation of our method on 3DFront and Matterport3D super-resolution and point cloud to surface reconstruction tasks in Fig.~\ref{fig:appendix_superresolution} and Fig.~\ref{fig:appendix_surface_reconstruction} respectively. Qualitative evaluation on ShapeNet for both of the tasks is provided in Fig.~\ref{fig:appendix_shapenet}. Further, additional qualitative visualization for \textit{Effect of retrieval and attention-based refinement} (main paper section~4.3) is provided in Fig.~\ref{fig:appendix_components}. 


\section{Additional Discussion}

The result in the main paper as well is the additional experiments in this document show the broad applicability of our method, achieving state-of-the-art reconstruction and super-resolution outputs.
Nevertheless, our approach still has limitations as discussed in the main paper.
In particular, if the retrieval approximations are suboptimal, they will not help in the refinement process.
Fig.~\ref{fig:bad_retrieval_3dfront} visualizes some samples where the retrieval approximations don't help the reconstruction.
However, in these cases, even though the retrievals don't help the reconstruction, they also don't worsen the reconstruction.
This is achieved by the blending network effectively ignoring the retrievals in such cases.
The dependence on good retrievals can be observed more clearly in the following experiment.
We train our retrieval and refinement networks on a ShapeNet subset of $8$ classes.
The dictionary is created using chunks from the same $8$ classes.
The trained networks are evaluated on a subset of new 5 classes.
As shown in Tab.~\ref{tab:appendix_unseen_classes} and Fig.~\ref{fig:shapenet_transfer_robust}, our method doesn't improve significantly over the backbone network due to low quality retrievals.
Compared to a naive fusion of features from retrievals however, which learns to rely on retrievals during training, our method is more robust.
A limitation of our method is cubic growth in number of chunks in the database with the decrease in patch size.
As observed in Tab. \ref{tab:patchsize_ablation}, smaller chunk retrievals help both retrieval and refinement.
This however comes at the cost of more patches in the database, making the database indexing and retrieval slower.

\end{appendix}

\end{document}